\pdfoutput=1
\documentclass[11pt]{article}

\usepackage{amsfonts}
\usepackage{amsmath}
\usepackage{amssymb}
\usepackage{amsthm}
\usepackage{appendix}
\usepackage{arydshln}
\usepackage{color}
\usepackage{dsfont}
\usepackage{empheq}
\usepackage{enumitem}
\usepackage[top= 1.3 in, bottom =1 in, left =0.75 in, right = 1 in]{geometry}
\usepackage{graphicx}
\usepackage{hyperref}
\hypersetup{colorlinks}
\usepackage[utf8]{inputenc}
\usepackage{listings}
\usepackage{placeins}
\usepackage{subcaption}
\usepackage{times}
\usepackage{url}

\usepackage[numbers]{natbib}
\usepackage{booktabs}
\usepackage[table]{xcolor}
\definecolor{lightgray}{gray}{0.9}
\let\oldtabular\tabular
\let\endoldtabular\endtabular
\renewenvironment{tabular}{\rowcolors{2}{lightgray}{white}\oldtabular}{\endoldtabular}

\title{IAIA-BL: A Case-based Interpretable Deep Learning Model for Classification of Mass Lesions in Digital Mammography}
\author{
Alina Jade Barnett \\ 
\rowcolor{white} \textit{Duke University}\\
\rowcolor{white} alina.barnett@duke.edu \\
\and 
Fides Regina Schwartz \\
\rowcolor{white} \textit{Duke University}\\
\rowcolor{white} Fides.Schwartz@duke.edu \\
\and 
Chaofan Tao \\ 
\rowcolor{white} \textit{Duke University}\\
\rowcolor{white} chaofan.tao@gmail.com \\
\and 
Chaofan Chen \\
\rowcolor{white} \textit{University of Maine}\\
\rowcolor{white} chaofan.chen@maine.edu \\
\and
Yinhao Ren \\
\rowcolor{white} \textit{Duke University}\\
\rowcolor{white} yinhao.ren@duke.edu \\
\and
Joseph Y. Lo \\
\rowcolor{white} \textit{Duke University}\\
\rowcolor{white} joseph.lo@duke.edu \\
\and 
Cynthia Rudin \\
\rowcolor{white} \textit{Duke University}\\
\rowcolor{white} cynthia@cs.duke.edu \\}

\date{}

\begin{document}

\maketitle

\begin{abstract}

Interpretability in machine learning models is important in high-stakes decisions, such as whether to order a biopsy based on a mammographic exam. Mammography poses important challenges that are not present in other computer vision tasks: datasets are small, confounding information is present, and it can be difficult even for a radiologist to decide between watchful waiting and biopsy based on a mammogram alone. In this work, we present a framework for interpretable machine learning-based mammography.
In addition to predicting whether a lesion is malignant or benign, our work aims to follow the reasoning processes of radiologists in detecting clinically relevant semantic features of each image, such as the characteristics of the mass margins. The framework includes a novel interpretable neural network algorithm that uses case-based reasoning for mammography. Our algorithm can incorporate a combination of data with whole image labelling and data with pixel-wise annotations, leading to better accuracy and interpretability even with a small number of images. Our interpretable models are able to highlight the classification-relevant parts of the image, whereas other methods highlight healthy tissue and confounding information. Our models are decision aids, rather than decision makers, aimed at better overall human-machine collaboration. We do not observe a loss in mass margin classification accuracy over a black box neural network trained on the same data. 
\end{abstract}

\section{Introduction}

\begin{figure}
        \begin{center}
        \includegraphics[width=.8\linewidth]{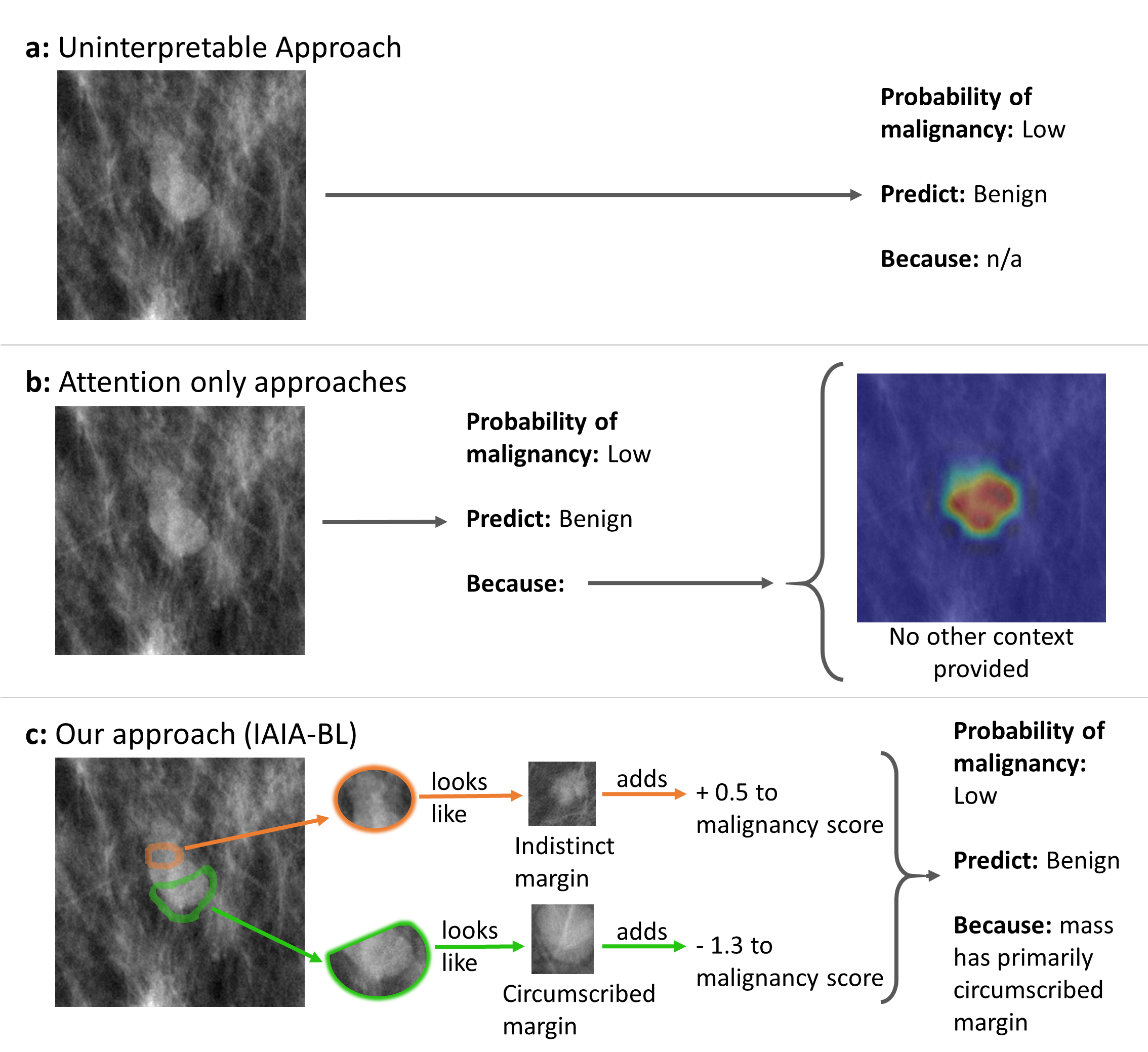}
        \end{center}
           \caption{(a) Uninterpretable approaches give no explanations for their output. (b) Other interpretable or explainable approaches might point out which regions are used for decision making, but provide no information about what attributes of the region are important for classification decisions. (c) IAIA-BL provides an explanation framework that localizes relevant areas, associates the relevant area with a specific medical feature and uses only the explained evidence to make a prediction. }
        \label{fig:intro_fig}
\end{figure}

AI is revolutionizing Radiology. Conventional machine learning is currently used for computer-aided detection (i.e., to detect ``lesion'' vs$.$ ``no lesion''), but to make a greater clinical contribution, future approaches need to be able to assist with harder tasks, such as ``should the patient get a biopsy for that lesion?'' Consider mammography, which aims to detect breast cancer, a leading cause of death in the USA \citep{cdc2019mortality}. In breast cancer screening, the majority of biopsies yield benign results, in the process subjecting many healthy patients to invasive testing and contribute to the societal cost of healthcare \citep{chhatwal2010optimal}; it is possible that machine learning might lead to improvements. 
In deciding how to treat a patient, radiologists must consider aspects of images that are so subtle that it is quite difficult for an untrained eye to identify even the important aspects of an image. These decisions can be challenging, even for the most experienced radiologists, as shown by relatively low inter-rater agreement scores for such difficult tasks \citep{park2007observer, abdullah2009breast, baker1996breast, lazarus2006bi, el2015breast}. To ensure clinical acceptance, an AI tool will need to provide its reasoning process to its human radiologist collaborators in order to be a useful aide in these difficult and high-stakes decision-making processes  \citep{edwards-fda, soffer2019convolutional}.

Despite the hope of computer-aided radiology for mammography, there are serious concerns with present methods, primarily, \textit{confounding}. Confounding occurs when the predictive model is using incorrect information or reasoning to make a decision, even if the decision is correct.  In previous studies, researchers created models that appeared to perform well on their test sets, yet upon further inspection, based their decisions on confounding information (e.g., type of equipment) rather than medical information \citep{badgeley2019deep,winkler2019association,zech2018variable}. If a model simply relied upon confounders such as the type of equipment used to take the mammogram or proxies for the patient's age such as the density of the breast tissue, the model would likely fail to generalize. This problem is exacerbated by the fact that there are few publicly available mammography datasets, so many models are trained on relatively few cases. 

Ideally, the reasoning process of any model would be similar to that of an actual radiologist, who looks at specific aspects of the image that are known to be important, based on the physiology of how lesions develop within breast tissue. If this reasoning process were correct, it would lead to a higher chance that: (1) The model could generalize beyond its finite training set. (2) The model's reasoning process could be useful information for doctors, even if its prediction is sometimes incorrect. (3) It would be much easier to troubleshoot or evaluate trustworthiness of the model, since it is not a black box. (4) The model's reasoning and reporting process could align with the structured lexicon that radiologists use to report results, such as the Breast Imaging-Reporting and Data System (BI-RADS) \citep{sickles2013acr} for breast cancer and other similar lexicons from the American College of Radiology.

Thus, unlike existing black-box systems that aim to replace a doctor \citep{mckinney2020international}, we aim to create an interpretable AI algorithm for breast lesions (abbreviated IAIA-BL) whose explicit reasoning can be understood and verified by a medical practitioner. Our novel deep learning architecture enables IAIA-BL to provide an explanation that shows the underlying decision-making process for each case. Figure \ref{fig:intro_fig}c shows an example of how this works: the algorithm highlights parts of the image, explains that it considers these parts of images similar to prototypical cases it has seen before, and provides a score for the probability of the specific diagnosis (breast mass with mostly circumscribed margin) for this image as well as the likelihood of malignancy. IAIA-BL provides the radiologist with the means not to simply trust the AI but to check its output for plausibility, and overrule it when necessary. 
As far as we know, ours is the first work that applies case-based reasoning using interpretable deep learning techniques to analyzing medical images. Our approach, with its inherently interpretable reasoning process, contrasts directly with previous work that relied on posthoc explanation techniques to explain a trained black-box model, or work that relies on attention mechanisms to highlight the parts of an input image upon which the model prediction is based. 
The reasoning process the network explains to the humans \textit{is} the reasoning process it is using to understand the image itself. 

As shown partly in Figure \ref{fig:intro_fig}c, the framework we establish aims to identify not just whether a lesion is malignant or benign, but aims to help with the full reasoning process in the decision of whether to perform a biopsy. Several prediction problems are involved in determining the mass margin and shape of a lesion, which a radiologist would need to consider as part of the required pipeline laid out by the American College of Radiology \citep{sickles2013acr}. In our framework, each of these prediction problems is handled by interpretable machine learning.

Machine learning becomes challenging with smaller datasets, such as those available for mammography. To extract more information from our dataset, we collected a small set of pixel-level (``fine'') annotations from our radiology team which permitted better generalization using a smaller number of images; that is, fine annotation on only 30 images of our limited data set (1136 annotated mammographic images from 484 patients with lesions) enabled high-quality reasoning and prediction. This novel approach can reduce the confounding in deep learning by leveraging both relatively-abundant coarsely-annotated data and a small amount of finely annotated data. Most fine-grained classification algorithms either assume the availability of fine-grained part labels for all input data and use them, or those fine-grained labels are completely ignored. Our approach provides a middle ground by using both data with and without fine-grained annotations, which takes full advantage of the information available. This approach is also practical in the sense that for real-world problems, annotated data are relatively less abundant and more expensive to obtain.

The main contributions of our paper are as follows: 
\begin{itemize}

    \item We developed the first inherently interpretable ML-based system for medical imaging that goes beyond simple attention in its explanations. Our system, IAIA-BL, makes predictions for mammographic breast masses by comparing test mammograms with \textit{prototypical} images of various mass margin types. 
    
    \item We developed a novel training scheme for our IAIA-BL which allows it to incorporate prior knowledge in the form of fine-grained expert image annotations. Using only a small number of finely annotated training data and imposing a novel fine-annotation loss on those data, IAIA-BL learns medically relevant prototypes, effectively addresses aspects of confounding issues in medical machine learning models, and sets our IAIA-BL apart from the ProtoPNet presented in \cite{PPNet} and other prior works.
    
    \item By changing the logic of ProtoPNet from max-pooling to top-k average pooling, we increase performance dramatically. This improvement can be used in any follow-on works that use ProtoPNet-style architecture.
    
    \item We developed a framework for machine learning-based mammography interpretation in line with the goals of radiologists: in addition to predicting whether a lesion is malignant or benign, our work aims to follow the reasoning processes of radiologists in detecting specific aspects of each image, such as the characteristics of the mass margins.
    
\end{itemize}

\section{Related Work}

\noindent\textbf{Background on Computer-Aided Detection/Diagnosis in Mammography.} Computer-aided detection systems flag suspicious lesions that may prompt the radiologist to recall a patient for additional imaging. Despite widespread clinical adoption, however, an influential study from 2015 found that current systems do not improve diagnoses in clinical practice \citep{lehman2015diagnostic}. More recent deep learning studies based on large numbers of cases have been reported to match or even exceed radiologist performance \citep{salim2020external,schaffter2020evaluation,mckinney2020international,wu2019deep, kim2020changes}. Going beyond lesion detection, computer-aided diagnosis systems provide additional diagnostic information such as to classify the lesion as benign vs$.$ malignant \citep{giger2008anniversary}. In this study, we seek to advance the underlying technology beyond that of previous computer-aided diagnosis approaches.

\begin{table}
\small
  \caption{We compare selected AI mammography techniques to our own.}
  \label{table:interp_compare}
  \centering
  \begin{tabular}{lcccc}
    \hline
     & Ours & \cite{wu2019deep} & \cite{kim2018icadx} & \cite{wu2018deepminer} \\
    \hline \hline
    Inherently interpretable model (not posthoc) & \checkmark &  \checkmark & \checkmark &  \\
    Provides global interpretability (on model) & \checkmark &  &  & \checkmark  \\
    Provides local interpretability (on each case) & \checkmark & \checkmark & \checkmark &  \\
    Explanation is guaranteed to match model reasoning & \checkmark &  &  &  \\
    Incorporate domain-specific terminology & \checkmark &  & \checkmark & \checkmark \\
    Provides similar prototypes for comparison & \checkmark &  &  &  \\
    Can incorporate fine annotation & \checkmark &  &  &  \\
    Can be trained on data with mixed labeling & \checkmark & \checkmark &  &  \\
    \hline
  \end{tabular}
\end{table}

\noindent\textbf{Background on Interpretable ML.}
In spite of their promising performance, deep neural networks are difficult to understand by humans. There are two distinct approaches to address this challenge: (1) \textit{Design inherently interpretable networks}, whose reasoning process is constrained to be understandable to humans. (2) \textit{Explain black box neural networks} posthoc by creating approximations, saliency maps, or derivatives. Posthoc explanations can be problematic; for instance \textit{saliency maps} highlight regions of the image, but can be unreliable and misleading, as they tend to highlight edges and do not show what computation is actually done with the highlighted pixels \citep{rudin2019stop,adebayo2018sanity,arun2020assessing}.
We avoid posthoc solutions in this work. There are several types of approaches in interpretable machine learning, including case-based reasoning (which we use here), forcing the network to use logical conditions within its last layers \citep[e.g.,][]{wu2019towards}, or disentangling the neural network's latent space \cite[e.g.,][]{chen2020concept}. Case-based reasoning models in medicine retrieve existing similar cases in order to determine how to handle a new case \citep{demigha2004case,macura1995macrad,floyd2000case,kobashi2006computer}.

IAIA-BL's framework incorporates the architecture of the inherently interpretable neural network \textit{ProtoPNet}, described in \cite{PPNet}. While ProtoPNet works well with bird classification, it was not able to be directly extended to mammograms, because of the problems with confounding, which is made worse by the dearth of data and the difficulty of the overall problem. IAIA-BL overcomes these obstacles through its framework, including incorporating fine-grained labels, modified modular training, and the addition of multi-stage reasoning wherein the model first determines the mass margin feature and uses that information to predict malignancy.

\noindent\textbf{Confounding and Fine Annotation.} \label{sec:confounding_RW}
Neural networks models often use context or confounding information instead of the information that a human would use to solve the same problem in both medical \citep{wang2019removing} and non-medical applications \citep{hu2020stratified}. The ability of these networks to use context or background information is so powerful that networks trained on images of \textit{only the background} outperform networks trained on images of \textit{only the object to be classified} \citep{dundar2017context,xiao2020noise}. 
For high-stakes applications in medicine, model decisions must use relevant medical information rather than context or background information. 
To address this we introduce an attention mechanism which redirects model attention to a selected part of the input image. Ways to direct model attention include data augmentation \citep{dundar2017context, LUO2016361, BMVC2016_110}, techniques that combat hand-selected confounders \citep{wang2019removing, tang2020mitigating}, techniques that combat learned confounders \citep{zhao2020training}, and an approach where a human critic manually approves the attention map during training \citep{Schramowski20}. Other techniques that show model attention but do not aim to change it are class activation maps \citep{zhou2016learning}, multi-attention CNNs \citep{zheng2017learning}, and recurrent attention CNNs \citep{fu2017look}. 
Our model uses an attention mechanism to incorporate expert annotations by adding a term to the objective function which penalizes attention outside of the regions marked as relevant by the radiologist-annotator. Mechanically, the method is most similar to that of \cite{tang2020mitigating}, but differs in that our class-specific attention mechanism asks for different attention from prototypes of different classes.

\section{Data and Methods}

\subsection{Framework}

Models need to be collaborators in the medical decision-making process in order to be useful. 
In mammography, the initial clinical decision is expressed as a BI-RADS category of 1 to 5, corresponding to the recommendation of whether the patient needs a biopsy \citep{orel1999bi,sickles2013acr}. An inscrutable model predicting malignant/benign is not useful as a decision aid, as a biopsy is recommended for every lesion with greater than 2\% chance of malignancy (BI-RADS 4 and 5). To alter clinical management, an interpretable model is needed to describe its reasoning process for why the patient should or should not receive a biopsy rather than provide an inscrutable prediction of malignancy.

Our AI approach includes an explicit reasoning system that resembles that of a practicing radiologist. Existing interpretability techniques for mammography include localization as in Figure \ref{fig:intro_fig}b, but there is no explanation of why an area is selected, what attributes of the region are used for classification, or what parts of the training set these associations are learned from. In a non-medical image analogy, though localization may provide a good interpretation for whether or not an image contains a vase (perhaps by highlighting the vase), it does not provide a good interpretation for classification of the vase pattern as Roman vs$.$ Asian antiquity (highlighting the vase pattern provides no further insight). Many recently published AI-mammography algorithms are still entirely uninterpretable as in Figure \ref{fig:intro_fig}a \citep{mckinney2020international}. 

We train a ProtoPNet-based IAIA-BL model for classifying mass margins of breast lesions -- such a model would be able to learn a set of prototypical features associated with each margin type, and predict the margin type of a previously unseen breast lesion based on the similarity of its margin with the learned features. It further uses the logits of the mass margin prediction to predict malignancy. Such a model may be integrated into a clinical support system for classifying breast lesions, because it can point to mammogram regions that resemble prototypical signs of cancerous growth (e.g., spiculated mass margin), and thereby assist doctors in making diagnoses. 

\subsection{Fine Annotations}

When starting to build an interpretable model for breast lesion classification, we na\"ively applied the case-based reasoning ProtoPNet architecture to medical images. Though the model appeared to be learning medically relevant features because of its high validation accuracy, the model made predictions using regions of the image that did not correspond to the medical information; in other words, the model used confounding information rather than medically relevant information. This is consistent with observations made by other groups of the dangers of confounding in medical imaging \citep{wang2019removing}. For non-medical image classification tasks, a typical approach might be to increase the size of the training set. However, as discussed above, one major barrier to implementation of AI in the medical field is the limited availability of annotated data \citep{soffer2019convolutional}. 

\begin{figure}
    \centering
        \begin{subfigure}{\textwidth}
        \centering
        \includegraphics[width=0.4\linewidth]{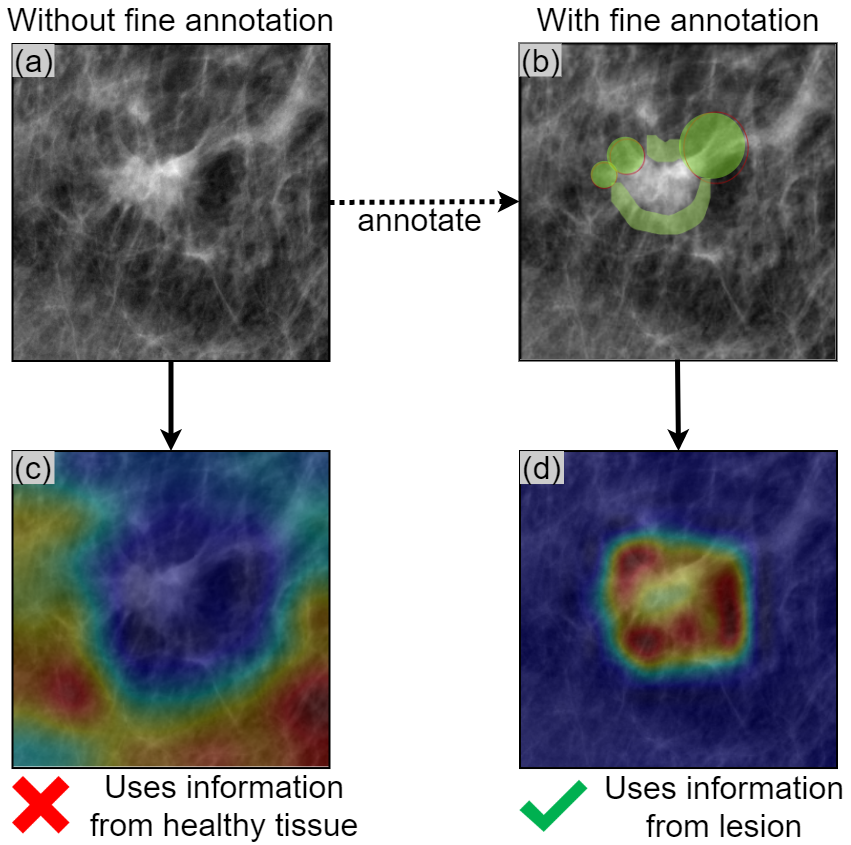}
        \caption{By introducing a constraint on model attention, we penalize the model for using confounding information. (a) Shows the lesion to be classified without further annotation. (b) The spicules of the lesion have been marked in green by a radiologist (circles for most prominent spicules, line for rest of lesion margin). (c) Without fine annotation in training, the activation map highlights confounding information. (d) With fine annotation in training, the activation map highlights relevant information -- areas that contain spicules. This is not penalized because the attention is within the area marked by the radiologist.} \label{fig:fine_attention}
        \end{subfigure}
        
        \begin{subfigure}{\textwidth}
        \centering
        \includegraphics[width=0.8\linewidth]{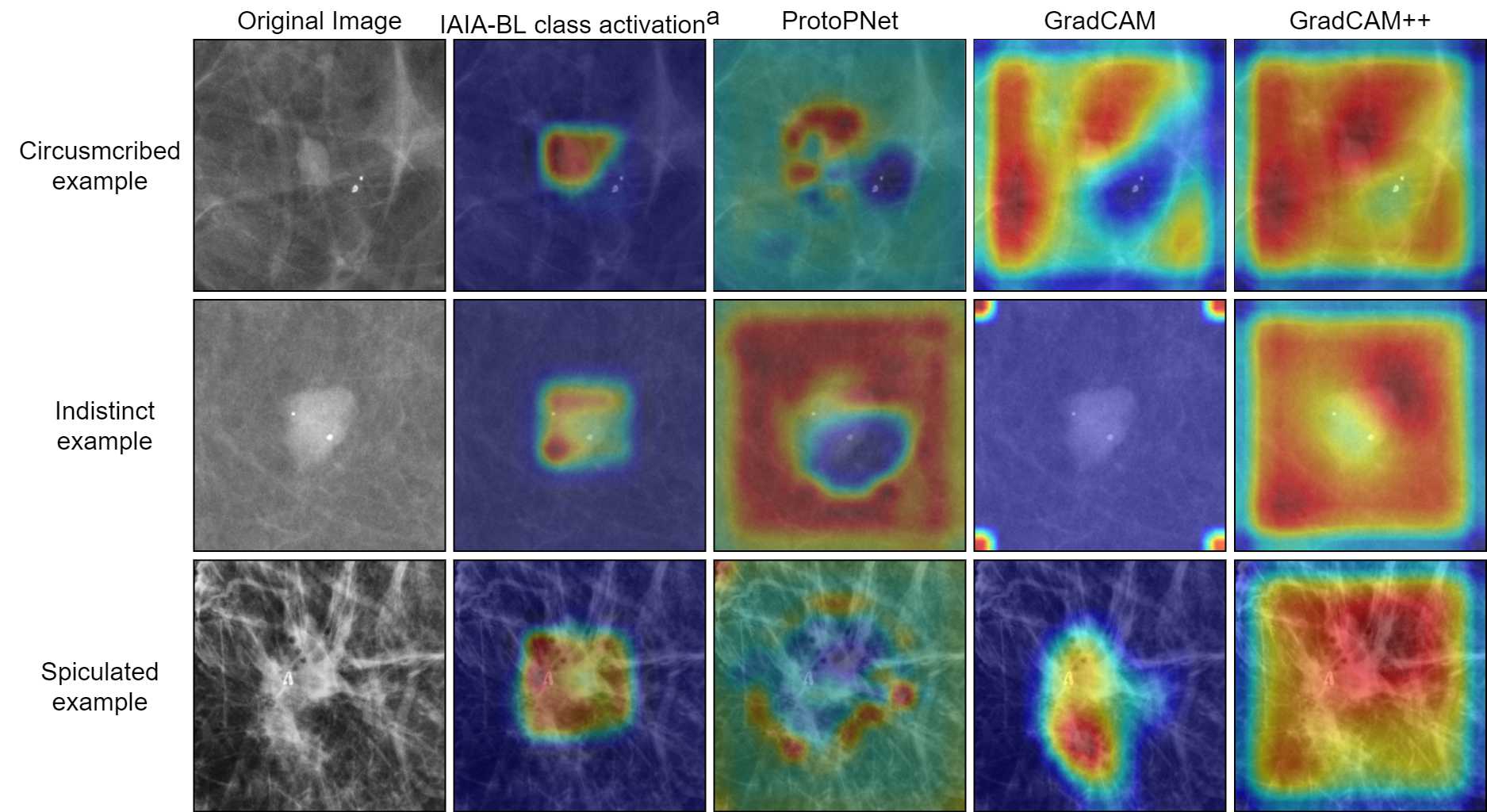}
        \caption{We compare explanations from two common saliency methods (GradCAM \cite{Selvaraju_2017_ICCV} and GradCAM++ \cite{chattopadhay2018grad}) to a class activation visualization derived from our method. The explanations from IAIA-BL are more likely to highlight the lesion and less likely to highlight the surrounding healthy tissue. This is shown quantitatively by the activation precision metric defined in Section \ref{subsec:activprec}. $^{\text{a}}$This single image visualization is a dramatic simplification of the full explanation that is generated by IAIA-BL. The IAIA-BL and ProtoPNet class activation visualizations shown in this figure are generated by taking the weighted average of prototype activation maps for all prototypes of the correct class (see Appendix \ref{app:compare_viz_expls}).} \label{fig:compare_gradcam_viz_to_us}
        \end{subfigure}
    \caption{IAIA-BL shows superior ability to highlight the medically relevant parts of a lesion.}
    \label{fig:fine_and_vis_compare}
\end{figure}

To make our limited institutional data stretch further, we designed a new training paradigm that incorporates additional expert annotation information on a subset of the existing patient cases. A radiologist (FS) annotated the area of a lesion image that indicates the mass margin for that lesion as in Figure \ref{fig:fine_attention}(b), with the most prominent and defining features marked by circles and the rest of the lesion margin highlighted by simple lines. The model incorporates the radiologist-supplied fine annotations by regularizing the activation of the prototypes over the image. It penalizes a prototype for activating anywhere on an image not of its class, or for activating  outside the region of the image marked ``relevant" by the radiologist. Figure \ref{fig:fine_attention}(c) shows an attention map that highlights confounding information and would be heavily penalized. Figure \ref{fig:fine_attention}(d) shows an attention map that highlights relevant information. By directing the network to the most relevant parts of the image, we set a strong prior on the network for where the useful information is centered in the image. Because these annotations are expensive to obtain, we designed the method to be able to use a small number of these finely annotated cases and a larger number of less expensive coarsely annotated cases. We include a training loss term in the objective as described in Section \ref{sec:model_training}. When using this, our performance and explanation both improve (see Appendix \ref{app:hyperparams}).

\subsection{IAIA-BL Model Architecture}

\begin{figure}[h]
        \begin{center}                        \includegraphics[width=\linewidth]{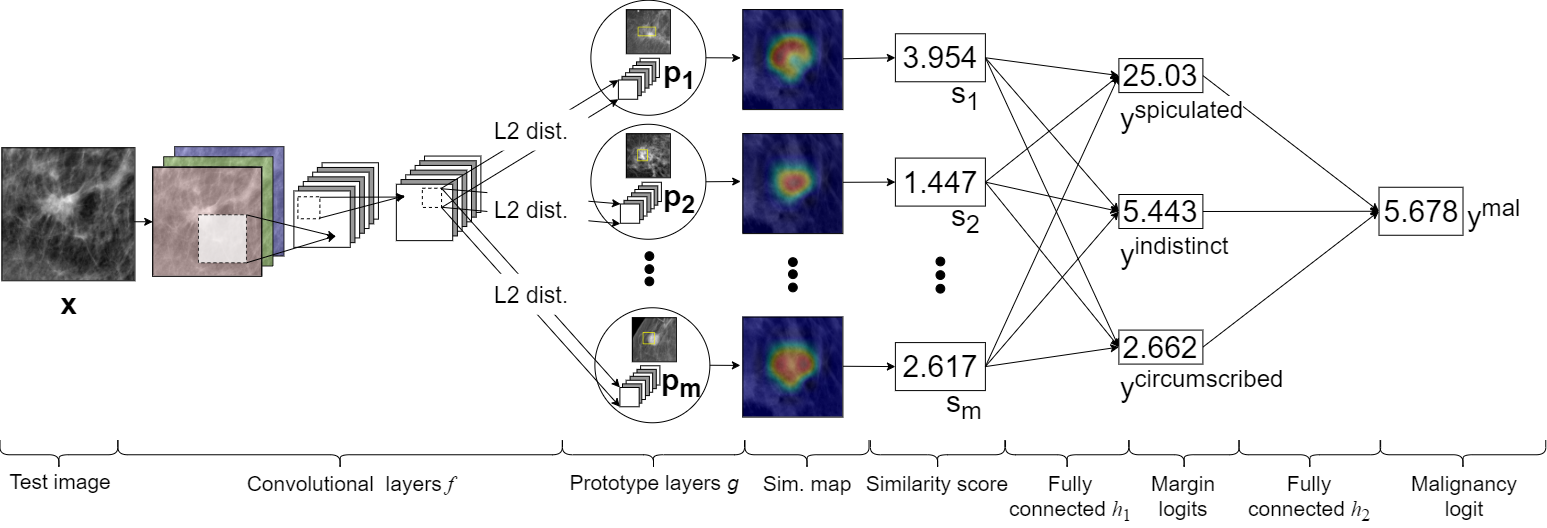}
        \end{center}
           \caption{The architecture of our prototype network. Test image $\textbf{x}$ feeds into convolutional layers $f$. Each patch of $f(\textbf{x})_l$ is compared to each learned prototype $\textbf{p}_\textbf{i}$ by calculating the $\ell_2$ distance between the patch and the prototype. The similarity map shows the closest (most ``activated," i.e., smallest $\ell_2$ distance) patches in red and the furthest patches in blue, overlaid on the test image. Similarity score $s_i$ is calculated from the corresponding similarity map. The similarity scores $\textbf{s}$ feed into fully connected layer $h_1$, outputting margin logits $\hat{\textbf{y}}^{\textrm{margin}}$. Margin logits $\hat{\textbf{y}}^{\textrm{margin}}$ feed into fully connected layer $h_2$, outputting malignancy logit ${y}^{\textrm{mal}}$.}
        \label{fig:architecture}
\end{figure}

We use a variation on the ProtoPNet architecture from \cite{PPNet} as the underlying architecture of our mass-margin classifier in IAIA-BL. 
Figure \ref{fig:architecture} gives an overview of our IAIA-BL model. Given a region of interest $\mathbf{x}$ in a mammogram, our IAIA-BL model first extracts useful features $f(\mathbf{x})$ for mass-margin classification, using a series of convolutional layers $f$ from a VGG-16 network \citep{simonyan2015very} pre-trained on ImageNet. Like \cite{PPNet}, our IAIA-BL model has a prototype layer $g$, which follows the convolutional layers $f$. The prototype layer $g$ contains $m$ prototypes $\mathbf{P}=\{\mathbf{p}_j\}_{j=1}^m$ learned from the training set. In our experiments, each prototype is a $1 \times 1$ patch with the same number (i.e., $512$) of channels as the convolutional feature maps $f(\mathbf{x})$. Since a prototype has the same number of channels but a smaller spatial dimension than the convolutional feature maps, we can interpret the prototype as representing a prototypical activation pattern of its class and we can visualize the prototype as a patch of the training image it appears in.
For example, our IAIA-BL model learns prototypical representations of spicules for the spiculated masses, fuzzy borders for indistinct masses, and clearly defined borders for circumscribed masses, and stores these prototypical representations as prototypes in the prototype layer $g$ for later comparison. 

Given an input image $\mathbf{x}$, the prototype layer $g$ compares the input image $\mathbf{x}$ with each prototype $\mathbf{p}_j$, by computing the squared $\ell_2$ distances $d_{j,l}$ between $\mathbf{p}_j$ and all $1 \times 1$ patches of the convolutional feature maps $f(\mathbf{x})$, and transforming the distances $d_{j,l}$ into similarity scores using:
\[
s_{j,l} = \log\frac{d_{j,l}+1}{d_{j,l}+\epsilon}
\]
where we have $d_{j,l} = \|\mathbf{p}_j-f(\mathbf{x})_l\|_2^2$, $l \in \{(1,1),...,(1,14),(2,1),...,(14,14)\}$ indexes the $1 \times 1$ patches of the $14 \times 14$ convolutional feature maps $f(\mathbf{x})$, and $f(\mathbf{x})_l$ is the $l$-th $1 \times 1$ patch of the convolutional feature maps $f(\mathbf{x})$. Conceptually, this means that if the input image $\mathbf{x}$ has spicules on the mass margin, its convolutional feature maps $f(\mathbf{x})$ will have patches $f(\mathbf{x})_l$ that represent spicules from the input image and are close (in $\ell_2$ distance in the latent space) to one or more prototypes $\mathbf{p}_j$ that are prototypical representations of spicules on the mass margin -- consequently, the similarity scores $s_{j,l}$ between those spiculated prototypes and those spiculated patches will be large. On the other hand, the patches of the convolutional feature maps corresponding to the image parts without spicules will have large $\ell_2$ distances from the prototypes representing spicules on the mass margin, and those patches will have small similarity scores with spiculated prototypes. Since each patch of the convolutional feature maps $f(\mathbf{x})$ has a similarity score with each prototype $\mathbf{p}_j$, the similarity scores between patches of $f(\mathbf{x})$ and a prototype $\mathbf{p}_j$ can be organized spatially into a similarity map, denoted $[s_{j,l}]_{l=(1,1)}^{(14,14)}$, which can then be upsampled to the size of the input image to produce a prototype activation map that identifies which parts of the input image are similar to the learned prototype. This is shown by the ``Sim. map'' column in Figure \ref{fig:architecture}.

Each similarity map $[s_{j,l}]_{l=(1,1)}^{(14,14)}$ between an input image and a prototype $\mathbf{p}_j$ is reduced to a single similarity score $s_j$, summarizing the degree of similarity between the input image and the learned prototype. Unlike \cite{PPNet} who used max-pooling to reduce each similarity map to a single similarity score, we used \textit{top-k average pooling} \citep[e.g., as in][]{kalchbrenner2014convolutional} because we found that our IAIA-BL trained with the relaxed cluster and separation costs outperforms the one trained with the original cluster and separation costs on the task of margin classification. For a given similarity map $[s_{j,l}]_{l=(1,1)}^{(14,14)}$, top-$k$ average pooling of the similarity map finds the $k$ highest similarity scores from the map and computes the average of those $k$ similarity scores. We denote this operation using $s_j = \text{AVGPOOL}({\text{topk}}([s_{j,l}]; k))$ in our paper. Note that max-pooling is a special case of top-$k$ average pooling, by using $k=1$. The top-$k$ average pooling allows the model to consider similarity between multiple parts of the input image and a mass-margin prototype, so that the similarity score after top-$k$ average pooling can be interpreted as how strong a prototypical feature is present (on average) in the $k$ most activated parts of the input image (instead of in the most activated part of the input image as in max-pooling). In the IAIA-BL, top-$5\%$ average pooling (i.e., $k=\lfloor5\%(14 \times 14)\rfloor=9$) is used to reduce each similarity map $[s_{j,l}]$ to a similarity score $s_j$. The similarity scores between an input image and the learned prototypes are illustrated in the ``Similarity score'' column in Figure \ref{fig:architecture}.

In IAIA-BL, we initially allocated $5$ prototypes for each of the mass-margin types represented in our dataset (circumscribed, indistinct, spiculated). The final IAIA-BL model presented has 4 prototypes for circumscribed mass margin, 3 prototypes for indistinct mass margin, and 4 prototypes for spiculated mass margin. We use $\text{class}(\mathbf{p}_j)$ to denote the class identity of a prototype.

Our IAIA-BL uses two fully connected layers. The first fully connected layer $h_1$ multiplies the vector of similarity scores $[s_1, ..., s_m]$ by a weight matrix to produce three output scores $\hat{y}^{\text{circumscribed}}$, $\hat{y}^{\text{indistinct}}$, and $\hat{y}^{\text{spiculated}}$, one for each margin type. These are (afterwards) normalized using a softmax function to generate the probabilities that the mass margin in the input image belongs to each of the three mass-margin types. The second fully connected layer $h_2$ then combines the vector of (unnormalized) mass-margin scores $\hat{\mathbf{y}}^{\text{margin}}=[\hat{y}^{\text{circumscribed}}, \hat{y}^{\text{indistinct}}, \hat{y}^{\text{spiculated}}]$ into a final score of malignancy $\hat{y}^{\text{mal}}$, which is passed into a logistic sigmoid function to produce a probability that the input image has a malignant breast cancer.

This architecture can provide both local interpretability by explaining each prediction in terms of the similarity between a given input image and the learned prototypes, as in Figure \ref{fig:IAIA_expls}, and global interpretability in terms of the clustering structure of the latent feature space (where semantically similar convolutional feature patches are clustered around prototypes representing the same semantic concepts). The set of learned prototypes is provided in Appendix \ref{app:prototype_sets}.

\subsection{Data}

Our dataset consists of 1136 digital screening mammogram images of masses in the breast from 484 patients at Duke University Health System. Each mass was coarsely annotated as a rectangular region of interest by one of four fellowship-trained breast imaging radiologists who had access to the original reports. For our training data, we cropped the region of interest as well as the surrounding area for context (rationale in Appendix \ref{app:context}). The BI-RADS features of mass shape and mass margin were labelled by one fellowship-trained breast imaging radiologist. The ground truth for malignancy of each mass is the result of definitive histopathology diagnosis.

The 1136 masses consisted of the following mass margins: 125 spiculated, 220 indistinct, 41 microlobulated, 579 obscured, and 171 circumscribed. We excluded lesions with microlobulated margins because of the small number of lesions represented. We excluded lesions with obscured margins because this margin class is not a good indicator for classifying a lesion as benign or malignant, but instead usually indicates the need for follow-up imaging. We split each remaining margin class into 73\% training, 12\% validation, and 15\% testing, ensuring no patient overlap between the testing set and other sets. All performances are based on the testing set alone (n=78). Given the small training set, we performed data augmentation such that each training image is randomly flipped, rotated, and undergoes random cropping with a crop size of 80\% of the image's original size. Each class is augmented to have 5000 images for the training set.

\subsection{Model Training} \label{sec:model_training}

The training of IAIA-BL differs from that of ProtoPNet \citep{PPNet} in three major ways: (1) IAIA-BL was trained with a fine-annotation loss which penalizes prototype activations on medically irrelevant regions for the subset of data with fine annotations. (2) IAIA-BL considers the top $5\%$ of the most activated convolutional patches that are closest to each prototype, instead of only the top most activated patch as in ProtoPNet (using the single max would be equivalent to using the top 0.5\%).
(3) We include an additional fully connected layer to transform mass margin scores $\hat{\mathbf{y}}^{\text{margin}}$ to malignancy score $y^{\text{mal}}$ whose training is isolated from the rest of the network. 

We represent the dataset of $n$ training images $\mathbf{x}_i$, with mass-margin labels $y^{\text{margin}}_i$ and malignancy labels $y_i^{\text{mal}}$, as $D=\{(\mathbf{x}_i, y^{\text{margin}}_i, y_i^{\text{mal}})\}_{i=1}^n$. A small subset $D' \subseteq D$ in the training set comes with fine annotations. For a training instance $(\mathbf{x}_i, y^{\text{margin}}_i, y_i^{\text{mal}}) \in D'$ that comes with the radiologist's (fine) annotations of where medically relevant information is in that training image, we define a fine-annotation mask $\textbf{m}_i$, such that $\textbf{m}_i$ takes the value $0$ at those pixels that are marked as ``relevant to mass margin identification,'' and takes the value $1$ at other pixels. Each fine-annotation mask $\textbf{m}_i$ has the same spatial dimensions (height and width) as the training image $\textbf{x}_i$.

The training of IAIA-BL is divided into four stages: (A1) training of the convolutional layers $f$ and the prototype layer $g$; (A2) projection of prototypes; (A3) training of the first fully connected layer $h_1$ for predicting mass-margin types; and (B) training of the second fully connected layer $h_2$ for predicting malignancy probability. Stages A1, A2, and A3 are repeated until the training loss for predicting mass-margin types converges, then we move to Stage B. By training Stage B after convergence for mass margin classification, we ensure that the mass margin classifier is not biased by the malignancy labels.
    
\textbf{Stage A1:} In the first training stage, we aim to learn meaningful convolutional features that can be clustered around prototypes that activate on medically relevant parts of a given patch. In particular, we want convolutional features that represent a particular mass-margin type to be clustered in latent space around a prototype of that particular mass-margin type, and to be far away from a prototype of other mass-margin types. As in \cite{PPNet}, we jointly optimize the parameters $\mathbf{\theta}_f$ of the convolutional layers $f$, and the prototypes $\mathbf{p}_1$, ..., $\mathbf{p}_m$ in the prototype layer $g$, while keeping the two fully connected layers $h_1$ and $h_2$ fixed. Differing from \cite{PPNet}, we minimize the following training loss:
\begin{align}
    \textrm{min}_{\mathbf{\theta}_f, \mathbf{p}_1, ..., \mathbf{p}_m} & \frac{1}{n} \sum_{i=1}^n \textrm{CrossEntropy}(h_1 \circ \textrm{AVGPOOL} \circ \textrm{topk} \circ g \circ f(\mathbf{x}_i), y^{\text{margin}}_i) \nonumber\\ +& \lambda_c \textrm{ClusterCost} + \lambda_s \textrm{SeparationCost} + \lambda_f \textrm{FineLoss}.
    \label{eq:joint_training_obj}
\end{align}
The cross-entropy loss in Equation (\ref{eq:joint_training_obj}) penalizes  misclassification of mass-margin types on the training data (using the subnetwork that excludes the second fully connected layer $h_2$). It also ensures that the learned convolutional features and the learned prototypes are relevant for predicting mass-margin types. Differing from Chen et al. \citep{PPNet} by the use of $\mathrm{mink}$ instead of $\min$, the cluster and separation cost are defined by:
\begin{align}
\textrm{ClusterCost} =& \frac{1}{n}\sum_{i=1}^{n} \min_{j:\text{class}(\mathbf{p}_j)=y^{\text{margin}}_i} \left(\frac{1}{k}\sum\mathrm{mink}_{\mathbf{z} \in \text{patches}(f(\mathbf{x}_i))}\left(\|\mathbf{z}-\mathbf{p}_j\|_2^2\right)\right), \label{eq:cluster_cost}\\
\textrm{SeparationCost} =& -\frac{1}{n}\sum_{i=1}^{n} \min_{j:\text{class}(\mathbf{p}_j) \neq y^{\text{margin}}_i} \left(\frac{1}{k}\sum\mathrm{mink}_{\mathbf{z} \in \text{patches}(f(\mathbf{x}_i))}\left(\|\mathbf{z}-\mathbf{p}_j\|_2^2\right)\right), \label{eq:separation_cost}
\end{align}
where $\mathrm{mink}$ gives the $k$ smallest squared distances between the convolutional patches of a training image and the $j$-th prototype, and $\frac{1}{k}\sum\mathrm{mink}$ gives the average squared distance over the $k$ smallest distances between the convolutional patches of a training image and the $j$-th prototype. The minimization of the above cluster cost encourages every training image to have $k$ convolutional feature patches that are close to a prototype of the same mass-margin type. The minimization of the separation cost increases the average of the $k$ smallest squared distances between the convolutional patches of a training image and a prototype not of the same class as the training image. This separates convolutional feature clusters of different mass-margin types. Note that the cluster and separation costs we used in IAIA-BL are relaxations of those used in \cite{PPNet}, in the sense that when we set $k = 1$, the above cluster and separation costs become the same as those used in \cite{PPNet}. Empirically, we found that IAIA-BL trained with the relaxed cluster and separation costs outperforms the one trained with the original (i.e., $k=1$) cluster and separation costs on the task of margin classification, possibly because the relaxed cluster and separation cost (along with the top-$k$ average pooling) allows the gradient of the loss function to back-propagate through $k$ convolutional patches, instead of just $1$ patch, during training -- consequently, the gradient will be \textit{less sensitive} and \textit{more robust} to changes in the location of the most activated convolutional patch by each prototype.

The fine-annotation loss (FineLoss) is entirely new to this paper. The purpose of the fine-annotation loss is to penalize prototype activations on medically irrelevant regions of radiologist-annotated training mammograms. The fine-annotation loss is defined by:
\begin{equation}
\textrm{FineLoss} = \sum_{i \in D'}  \left(\sum_{j: \text{class}(\mathbf{p}_j) = y^{\text{margin}}_i} \|\mathbf{m}_i \odot \textrm{Upsample}(g_{\mathbf{p}_j}(f(\mathbf{x}_i)))\|_2
+ \sum_{j: \text{class}(\mathbf{p}_j) \neq y^{\text{margin}}_i} \|g_{\mathbf{p}_j}(f(\mathbf{x}_i))\|_2\right)
\end{equation}
where $g_{\mathbf{p}_j}(f(\mathbf{x}_i))$ computes the similarity map between patches of the convolutional features $f(\mathbf{x}_i)$ and the $j$-th prototype $\mathbf{p}_j$, and $\textrm{Upsample}(g_{\mathbf{p}_j}(f(\mathbf{x}_i)))$ computes bilinear upsampling of the similarity map $g_{\mathbf{p}_j}(f(\mathbf{x}_i))$ to yield a prototype activation map of the same dimensions (height and width) as the fine-annotation mask. 

Since the fine-annotation mask $\mathbf{m}_i$ and the prototype activation map (denoted $\textrm{Upsample}(g_{\mathbf{p}_j}(f(\mathbf{x}_i)))$) have the same dimensions, we can compute a Hadamard (component-wise) product between them. For a given training instance $i \in D'$ that comes with a fine-annotation mask $\mathbf{m}_i$ and has a mass-margin type $y^{\text{margin}}_i$, and for a mass-margin prototype $\mathbf{p}_j$ with $\text{class}(\mathbf{p}_j) = y^{\text{margin}}_i$, the Hadamard product between $\mathbf{m}_i$ and the prototype activation map (denoted $\textrm{Upsample}(g_{\mathbf{p}_j}(f(\mathbf{x}_i)))$) gives a map that shows the prototype activations in the medically irrelevant regions of the training image (because the fine-annotation mask $\mathbf{m}_i$ takes the value $1$ at medically irrelevant pixels, and $0$ at medically relevant pixels). Hence, the first sum in the parentheses of the fine-annotation loss tends to reduce the amount of prototype activations in medically irrelevant regions when those prototypes are of the same class as the training image $i \in D'$. This, in turn, reinforces the training algorithm to learn prototypes that encode medically relevant mass-margin features for the prototypes' designated classes. On the other hand, for a given training instance $i \in D'$, the second sum in the parentheses of the fine-annotation loss penalizes any amount of prototype activation for a mass-margin prototype $\mathbf{p}_j$ with $\text{class}(\mathbf{p}_j) \neq y^{\text{margin}}_i$. This promotes the learning of prototypes that stay away from any features that could appear in classes that are not the prototypes' designated classes, so that the prototypes of a particular class represent distinguishing features of that class.

To incorporate the training data with fine annotations into model training, we optimize the convolutional layers $f$ and the prototype layer $g$ by minimizing the training objective in Equation (\ref{eq:joint_training_obj}), using stochastic gradient descent with $75$ training examples with lesion-scale annotation and $10$ training examples with fine annotations. The fine-annotation loss on a lesion-scale annotation penalizes activation outside of the area marked as the lesion, whereas the fine-annotation loss on a finely annotated image penalizes activation outside of the region ``relevant to the mass margin class'' as marked by the radiologist. 

The prototype layer was initialized randomly using the uniform distribution over a unit hypercube (because the convolutional features from the last convolutional layer all lie between $0$ and $1$). 

\textbf{Stage A2:} As in \cite{PPNet}, we project the prototypes $\textbf{p}_j$ onto the nearest convolutional feature patch from the training set $D$, of the same class as $\textbf{p}_j$. See \cite{PPNet} for a detailed description of how a prototype is visualized.

\textbf{Stage A3:} 
After the previous two training stages, the (medically relevant) convolutional features have been clustered in latent space around mass-margin prototypes that are identical to some (medically relevant) convolutional features from training images, which can be visualized in the original image space. In this stage, we fine-tune the first fully connected layer $h_1$ to further increase the accuracy in predicting mass-margin types. 
In particular, we fix the parameters $\mathbf{\theta}_f$ of the convolutional layers $f$ and the prototypes $\mathbf{p}_1$, ..., $\mathbf{p}_m$, and minimize the following training objective with respect to the parameters $\mathbf{\theta}_{h_1}$ of the first fully connected layer $h_1$:
\begin{equation}
\textrm{min}_{\mathbf{\theta}_{h_1}} \frac{1}{n} \sum_{i=1}^n \textrm{CrossEntropy}(h_1 \circ \textrm{AVGPOOL} \circ \textrm{topk} \circ g \circ f(\mathbf{x}_i), y^{\text{margin}}_i). \label{eq:h1_training}
\end{equation}

The first time we enter stage A3, we initialize connections in fully connected layer $h_1$ to a value of 1 for prototypes that are positive for that mass margin, -1 otherwise.

\textbf{Stage B:} In this stage, we train the second fully connected layer $h_2$ for predicting malignancy probability, using a logistic regression model whose input is the (unnormalized) mass-margin scores produced by the first fully connect layer $h_1$, and whose output is the probability of malignancy. To prevent the malignancy information from biasing the mass margin classification, we train the model in a modular style and it is not trained completely end-to-end in any stage, i.e., there is no return to Stage A from Stage B.

\section{Experiments and Results} \label{sec:exps_and_results}

\subsection{Performance Metrics}

We use the AUROC (area under receiver operator characteristic curve) for each of the three mass margin classes as the performance metric for both mass-margin prediction and malignancy prediction. An image-weighted average of these AUROCs to measures overall performance. 95\% confidence intervals were derived using Delong's method \citep{delong1988comparing, sun2014fast}.

Cohen $\kappa$ shows the agreement between our model’s predictions and the physician-annotator's labels for the mass margin prediction task. We use Cohen $\kappa$ to compare our model's agreement to the agreement of physicians with each other from previous studies \citep{park2007observer, abdullah2009breast, baker1996breast, RAWASHDEH2018294, lazarus2006bi}. 95\% confidence intervals were derived using non-parametric bootstrap resampling with 5000 samples each equal to the size of the test set.

\subsection{Interpretability Metric}\label{subsec:activprec}

We designed the interpretability metric \textit{activation precision} to quantify what proportion of the information used to classify the mass margin comes from the relevant region as marked by the radiologist-annotator. Using the notations defined in Section \ref{sec:model_training}, the activation precision for a single prototype $\mathbf{p}_j$ on a single image $\mathbf{x}_i$ that has mass-margin type $y^{\text{margin}}_i$ and comes with a fine-annotation mask $\mathbf{m}_i$, is defined as:
\begin{equation}\label{eq:ap}
     \textrm{AP}(\mathbf{p}_j, \mathbf{x}_i, y^{\text{margin}}_i, \mathbf{m}_i) = \left(\frac{\sum \left[(1 - \mathbf{m}_i) \odot T_{\tau}\left(\textrm{Upsample}\left(g_{\mathbf{p}_j}(f(\mathbf{x}_i))\right)\right)\right]}{\sum T_{\tau}\left(\textrm{Upsample}\left(g_{\mathbf{p}_j}(f(\mathbf{x}_i))\right)\right)}\right)\text{ where } \text{class}(\mathbf{p}_j)=y^{\text{margin}}_i,
\end{equation}
where $T_{\tau}$ is a threshold function that returns the top $(1-\tau) \times 100\%$ of the input values as $1$ and the bottom $\tau \times 100\%$ as $0$. Activation precision is only defined where the prototype has the same class identity as the image. The fraction in Equation (\ref{eq:ap}) gives a proportion of highly activated pixels that are medically relevant. We elaborate on this in Appendix \ref{app:activation_precision}. To evaluate activation precision for GradCAM \cite{Selvaraju_2017_ICCV} and GradCAM++ \cite{chattopadhay2018grad}, we calculate as in Equation (\ref{eq:ap}) but replace the prototype activation map $\textrm{Upsample}\left(g_{\mathbf{p}_j}(f(\mathbf{x}_i))\right)$ with the normalized gradient map for the correct class.

Note that we do not compute the proportion of medically relevant pixels that are highly activated, i.e., the denominator in Equation (\ref{eq:ap}) is \textit{not} the number of medically relevant pixels (given by $\sum (1 - \mathbf{m}_i)$). This is because we do not require each prototype to detect the entire mass margin (that was annotated by a doctor), but rather, we expect each prototype to detect a differentiating feature that may only be present at \textit{parts} of a mass margin. Since a prototype may only focus on parts of a margin, intersection over union or measuring the proportion of medically relevant pixels on which the prototype activates highly would not be appropriate metrics.

We can extend the above definition to a dataset $D'$, as:
\begin{equation}
    \textrm{AP}(\mathbf{p}_j, D') = \frac{1}{|D'|}\sum_{i \in D'} \textrm{AP}(\mathbf{p}_j, \mathbf{x}_i, y^{\text{margin}}_i, \mathbf{m}_i),
\end{equation}
and further to a set of prototypes $\mathbf{P} = \{\mathbf{p}_j\}_{j=1}^m$:
\begin{equation}
    \textrm{AP}(\mathbf{P}, D') = \frac{1}{|D'|m}\sum_{i \in D'}\sum_{j=1}^m \textrm{AP}(\mathbf{p}_j, \mathbf{x}_i, y^{\text{margin}}_i, \mathbf{m}_i).
\end{equation}

Activation precision is a measure of interpretability, in the sense that the higher the activation precision, the better a prototype (or a set of prototypes) is at detecting medically relevant features for mass-margin classification. In our experiments, we used $\tau = 0.95$ because IAIA-BL uses the top 5\% of activated patches in its predictions. 95\% confidence intervals were derived using non-parametric bootstrap resampling with 5000 samples each equal to the size of the test set.

Activation precision can be measured both at \textit{lesion-scale} (i.e., is the activation within the lesion area and not the added context window?) and at \textit{fine-scale} (i.e., is the activation on the specific part of the margin marked relevant by the radiologist?).

\subsection{Mass Margin Prediction}
\label{sec:results_margin}

We compare the following models.

    \textbf{IAIA-BL.} We used ProtoPNet with VGG-16 pre-trained on ImageNet as the base architecture trained for 50 epochs because model training converges between 40 and 50 epochs. The final model is trained on the combination of the training set and validation set, and tested on a test set never before seen in training. See Appendix \ref{app:hyperparams} for hyperparameters. Our model can be fully trained on one P100 GPU in 50 hours.

    \textbf{Baseline 1: Original ProtoPNet from \citep{PPNet}.} The original ProtoPNet architecture does not use fine annotation loss, and uses max pooling logic where IAIA-BL uses top-$k$ average pooling logic. This change is equivalent to changing all uses of $\textrm{AVGPOOL}$ to $\max$ in Equations (\ref{eq:joint_training_obj}) and (\ref{eq:h1_training}); and changing $\mathrm{mink}$ to $\min$ in Equations (\ref{eq:cluster_cost}) and (\ref{eq:separation_cost}). 

    \textbf{Baselines 2a and 2b: VGG-16 \citep{simonyan2015very} with GradCAM \cite{Selvaraju_2017_ICCV} and GradCAM++ \cite{chattopadhay2018grad}.} We trained a VGG-16 model with two added fully connected layers to account for the larger number of parameters in our model. Pre-trained on ImageNet, it was trained for 250 epochs and the epoch with the highest test accuracy is selected for comparison. There is no native way to incorporate our fine annotation into VGG-16. VGG-16 provides no inherent interpretability or localization. Using the posthoc GradCAM and GradCAM++ techniques we show localization information and calculate activation precision.

\begin{figure}
    \centering
        \begin{subfigure}{\textwidth}
        \includegraphics[width=\textwidth]{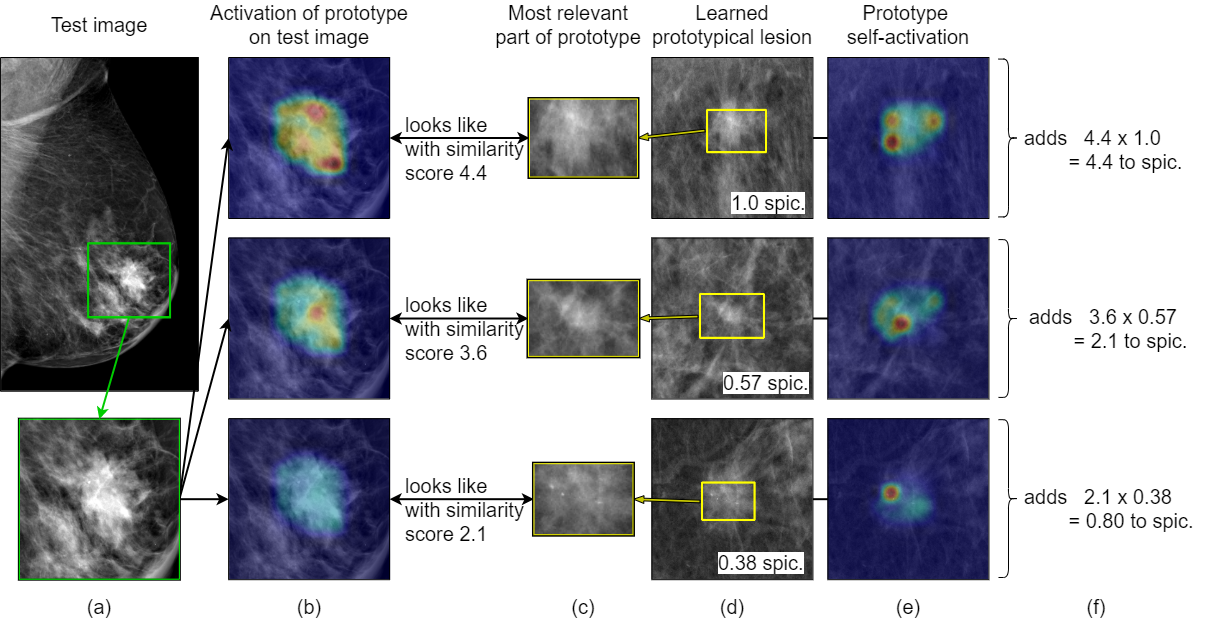}
        \caption{This spiculated lesion is correctly classified as spiculated.}
        \end{subfigure}
        
        \begin{subfigure}{\textwidth}
        \includegraphics[width=\textwidth]{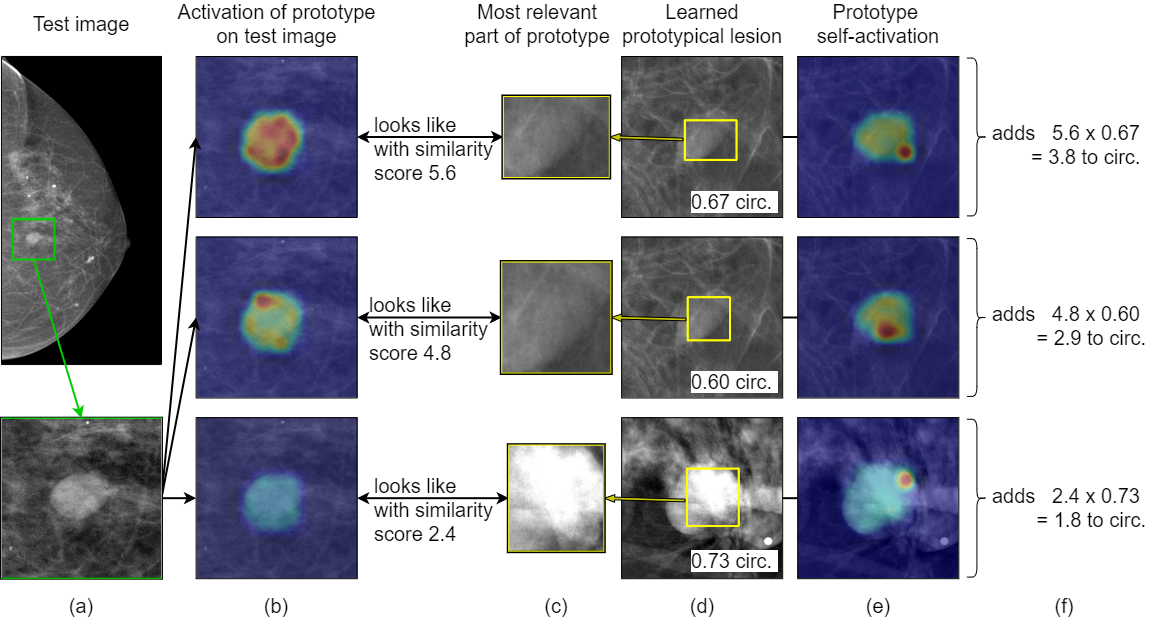}
        \caption{This circumscribed lesion is correctly classified as circumscribed.}
        \end{subfigure}
        
    \caption{Prediction explanations generated by IAIA-BL.}
    \label{fig:IAIA_expls}
\end{figure}

\begin{figure}
    \centering
        \begin{subfigure}{0.48\textwidth}
        \includegraphics[width=\textwidth]{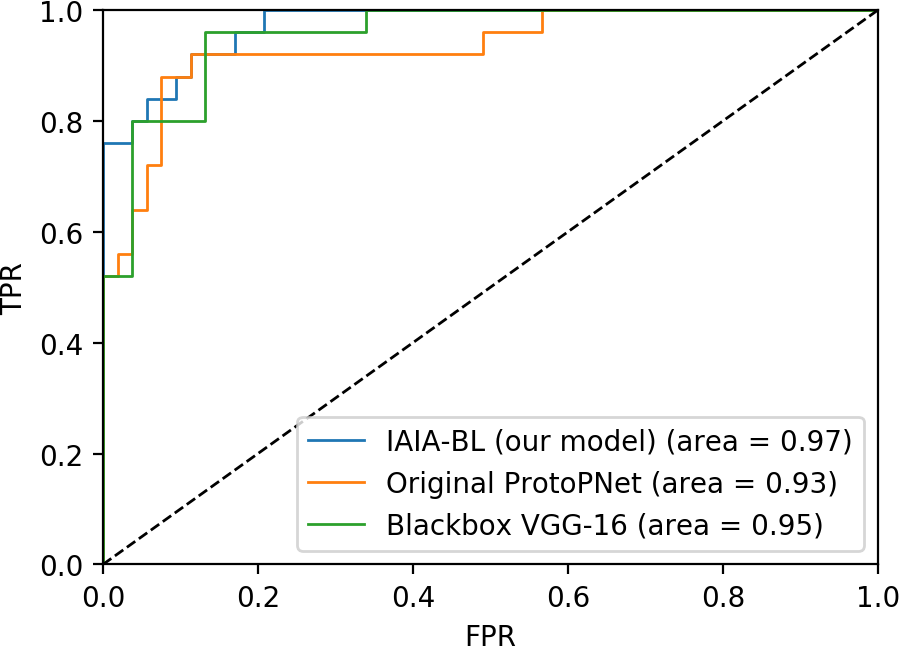}
        \caption{Circumscribed class.}
        \end{subfigure}
        \hfil
        \begin{subfigure}{0.48\textwidth}
        \includegraphics[width=\textwidth]{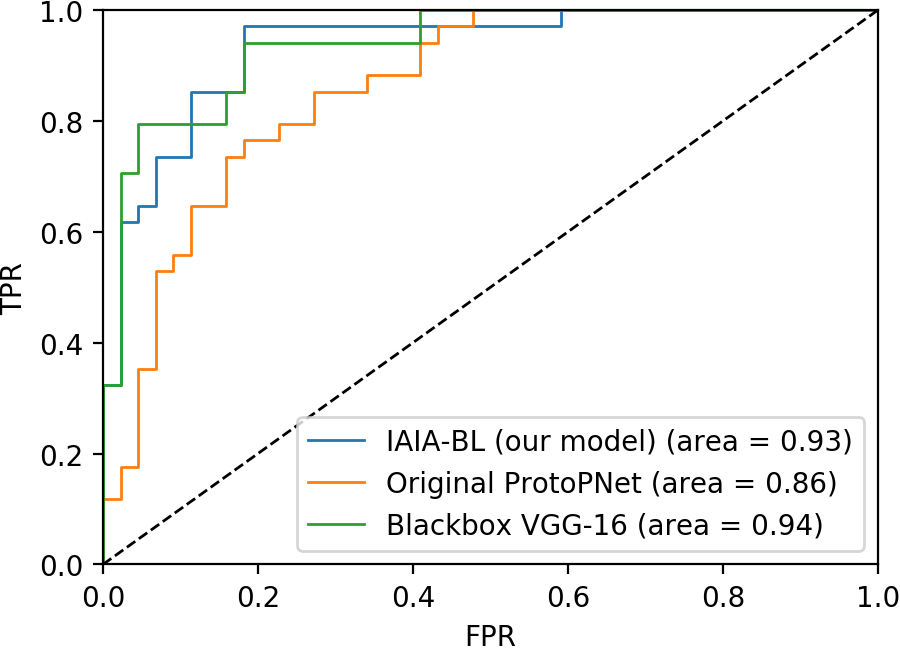}
        \caption{Indistinct class.}
        \end{subfigure}
        
        \begin{subfigure}{0.48\textwidth}
        \includegraphics[width=\textwidth]{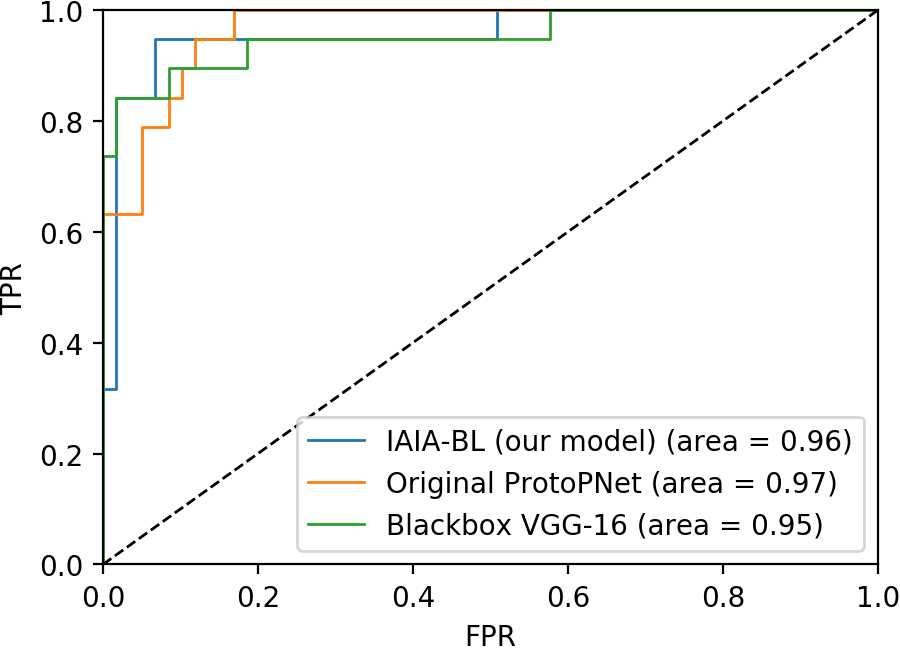}
        \caption{Spiculated class.}
        \end{subfigure}
        \hfil
        \begin{subfigure}{0.48\textwidth}
        \includegraphics[width=\textwidth]{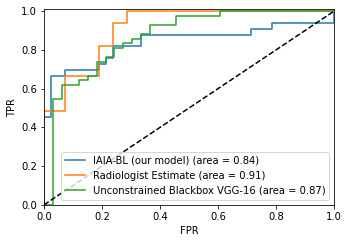}
        \caption{Malignancy.}
        \end{subfigure}
        
    \caption{ROC curves of our pruned model compared to unpruned baselines.}
    \label{fig:roc_massmargin_results}
\end{figure}

\setlength\tabcolsep{5 pt}
\begin{table}[ht]
\small
  \caption{Mass margin classification test results. The first five rows measure prediction performance, whereas the lower two rows measure interpretability performance. \textit{IAIA-BL:} The pruned IAIA-BL model as described in Section \ref{sec:exps_and_results}. \textit{ProtoPNet:} Original ProtoPNet. \textit{VGG-16 with GradCAM and GradCAM++:} Well-performing black box model VGG-16. The table shows that IAIA-BL's test AUROC performance with respect to all tasks is approximately as good as the best of the baselines. IAIA-BL's main advantage (interpretability) is shown in the bottom two rows of the table, where there is a huge drop in fine activation precision for original ProtoPNet and VGG-16 as compared with IAIA-BL. VGG-16 has no inherent interpretability but posthoc GradCAM and GradCAM++ provide localization information on which we measure activation precision. For each row, the best value is in \textbf{bold}, and values not significantly different than the best are in \textit{italics}. $^{\textrm{a}}$ Because this technique is posthoc, there is no guarantee that the generated explanation matches the model's decision making.
  }
  \label{tab:margin_results}
  \centering
  \begin{tabular}{llllll}
    \hline
    \rowcolor{white} & \multicolumn{4}{c}{Model} \\
    \rowcolor{white} &  &  & VGG-16 \cite{simonyan2015very}& VGG-16 \cite{simonyan2015very} \\
    \rowcolor{white} & IAIA-BL  & ProtoPNet \cite{PPNet} & with GradCAM \cite{Selvaraju_2017_ICCV} & with GradCAM++ \cite{chattopadhay2018grad} \\
    \hline
    Performance (AUROC) &  &  &  & \\
    ~~Mass Margin Class. & \textbf{0.951} [0.905, 0.996]  & \textit{0.911} [0.848, 0.974] & \textit{0.947} [0.898, 0.996] & \textit{0.947} [0.898, 0.996] \\
    ~~~~Spiculated vs. all & \textit{0.96} [0.90, 1.00] & \textbf{0.97} [0.93, 1.00] & \textit{0.95} [0.89, 1.00] & \textit{0.95} [0.89, 1.00] \\
    ~~~~Indistinct vs. all & \textit{0.93} [0.88, 0.99] & \textit{0.87} [0.78, 0.94] & \textbf{0.94} [0.89, 0.99] & \textbf{0.94} [0.89, 0.99] \\
    ~~~~Circumscribed vs. all & \textbf{0.97} [0.94, 1.00] & \textit{0.93} [0.87, 1.00] & \textit{0.95} [0.91, 1.00] & \textit{0.95} [0.91, 1.00]\\
    Cohen's $\kappa$ & \textbf{0.74} [0.60, 0.86] & \textit{0.64} [0.49, 0.78] & \textbf{0.74} [0.60, 0.87] & \textbf{0.74} [0.60, 0.87]\\
\hline
    Interpretability &  &    & &  \\
    ~~Fine-scale Act. Prec. & \textbf{0.41} [0.39, 0.45] & 0.24 [0.17, 0.31] & 0.21 [0.05, 0.43]$^{\textrm{a}}$ & 0.24 [0.08, 0.45]$^{\textrm{a}}$ \\
    ~~Lesion-scale Act. Prec. &  \textbf{0.94} [0.92, 0.97] & 0.51 [0.34, 0.68] & 0.45 [0.37, 0.54]$^{\textrm{a}}$ & 0.53 [0.44, 0.61]$^{\textrm{a}}$ \\
    \hline
  \end{tabular}
\end{table}

\textbf{Prediction Results:} 	
    Treating the radiologist annotations as the ground truth, pruned IAIA-BL achieves AUROCs as reported in Table \ref{tab:margin_results} and an accuracy for the overall mass margin classification task of 83\% (n=78, 95\% CI: 0.75\%, 0.92\%). (Without pruning, there is a 0.004 increase in AUROC for mass margin prediction.) Figure \ref{fig:roc_massmargin_results} shows ROC curves for all prediction tasks and all methods.

	Though there are many papers on computer vision with applications to mammography, few papers attempt to classify masses by margin type. We found only one study, \cite{kim2018icadx} who report an accuracy of mass margin prediction at 70.6\% and include more margin types than we do, but their provided results are not separated into different margin classes so we cannot directly compare. Further reducing comparability, that study used digitized mammography images from the DDSM database. Their model is not publicly available.
	
	Performance of IAIA-BL is better than that of ProtoPNet (Baseline 1), which does not have the stabilization of the gradient provided by the average pooling improvement of IAIA-BL. VGG-16 (Baselines 2a and 2b) performed comparably to IAIA-BL for AUROC. Remember that the baseline models are permitted to use confounding information that IAIA-BL is not encouraged to use, and we will see that when we consider the interpretability results. As we know, it is easy to perform well on training data despite using logic that a radiologist would claim is incorrect \citep{zech2018variable}.
	
    Another measure of performance we calculated is the Cohen $\kappa$ agreement between IAIA-BL and our human mass margin annotator on the test set. 
	We found ``substantial'' agreement with a $\kappa$ value of 0.74 (n=78, 95\% CI: 0.60, 0.86) \citep{landis1977application}, further broken down into circumscribed at 0.76 (95\% CI: 0.58, 0.90), indistinct at 0.69 (95\% CI: 0.51, 0.84), and spiculated at 0.78 (95\% CI: 0.61, 0.93). For this task of characterizing the mass margin in mammography, our performances were higher than the interobserver agreement of radiologists with each other, e.g., 0.61-0.65 in \cite{baker1996breast}, 0.58 in \cite{RAWASHDEH2018294}, and 0.48 in \cite{lazarus2006bi}. 
    
\textbf{Interpretability Results:} 
    To measure interpretability, we used the interpretability metric activation precision from Section \ref{subsec:activprec}, shown in the lower two rows of Table \ref{tab:margin_results}. For the unpruned IAIA-BL model (not shown in the table, because it is almost identical to IAIA-BL), the lesion-scale activation precision of the learned prototypes is 0.93 (95\% CI: 0.91, 0.96) and the fine-scale activation precision of the learned prototypes is 0.41 (95\% CI: 0.39, 0.43).
    
    Both ProtoPNet (Baseline 1) and VGG-16 (Baselines 2a and 2b) show lower activation precision than IAIA-BL. Both use information from image regions entirely outside the region that contains the lesion. The baseline models are not restricted from using confounding information, and thus do so freely. These models should not be used in practice for this reason. A visual comparison of activation maps (defined in Appendix \ref{app:compare_viz_expls}) is shown in Figure \ref{fig:compare_gradcam_viz_to_us}.

	\textbf{\textit{To summarize, IAIA-BL's predictive performance was as good or better than the analogous black-box model. Its performance in mimicking our annotator was better than the typical interannotator agreement between radiologists. 
	Its interpretability, measured by how well its attention agreed with a radiologist annotator's hand-drawn attention maps, exceeded that of existing methods and does not resort to post-hoc analysis.}}

\subsection{Malignancy Prediction}
\label{sec:results_mal}

Even though IAIA-BL is constrained to using only the results of the mass margin outputs to predict malignancy (rather than extra information that may be contained within the raw pixels of the image), IAIA-BL predicts mass malignancy with AUROC of 0.84 (n=75, 95\% CI: 0.74, 0.94). These results are interpretable in that they only use the mass margin scores to make their predictions. In the equation below, ${y}_i^{\text{circumscribed}}, {y}_i^{\text{indistinct}}, {y}_i^{\text{spiculated}}$ correspond to raw, unnormalized mass margin scores for circumscribed, indistinct and spiculated margin respectively. The conversion between mass margin scores and malignancy score ${y}_i^{\text{mal}}$ is the following concise linear model:
\begin{align}
    {y}_i^{\text{mal}} =& -16~ {y}_i^{\text{circumscribed}} -10 ~{y}_i^{\text{indistinct}} + 6~ {y}_i^{\text{spiculated}}, \text{ with} \\
    \text{Prob}(\text{malignancy}) =& \sigma(({y}_i^{\text{mal}} - 155) / 100 ), \text{ where } \sigma(t) = \frac{1}{1+\exp(-t)} \text{ is the logistic sigmoid function}.
\end{align}
As expected, a high spiculated score results in a high probability of malignancy, while high circumscribed or indistinct margin scores indicate a benign lesion. Each mass margin score is explained as in Figure \ref{fig:IAIA_expls}.

We remark that the prediction of whether a mass has 5\% or 95\% probability of being malignant would not alter the clinical management, since all lesions with \textgreater 2\% probability of malignancy would be recommended to undergo breast biopsy.

There are a variety of malignancy performance values reported in the literature, though not necessarily from the same population as ours, which means the results are not directly comparable. Some studies have reported better performance in predicting malignancy from BI-RADS features
\citep{elter2007prediction,benndorf2015external,burnside2009probabilistic}. 
If our dataset were larger, and if we had non-imaging features such as patient age, it could potentially boost performance.

\textbf{Baseline 3: Radiologist estimate.} During data collection, we asked radiologists to estimate the probability that the lesion will be malignant. There are several caveats for this estimate: radiologists do not perform this task in standard practice, instead they only provide a categorical recommendation for biopsy; the annotations were completed as part of a separate study that used consumer-grade monitors without the necessary specifications or calibrations of medical-grade displays. Nonetheless, these estimates represent the radiologist's ``best guess'' when given even more information than the model is provided.
The radiologists predicted mass malignancy on the test set with AUROC of 0.91 (n=75, 95\% CI: 0.85, 0.97). These radiologists are from Duke Hospital, and thus represent an extremely high quality of care for patients. \textbf{\textit{Using this as a reference standard, IAIA-BL is approximately 7\% in AUC away from the physicians.}} 

\textbf{Baseline 4: Unrestricted end-to-end VGG-16.} The uninterpretable VGG-16 baseline given the same image data, but not restricted to predicting on only mass margin results, achieves an AUROC of 0.87 (n=75, 95\% CI: 0.82, 0.93). Again, it is possible that VGG-16 uses confounding information; e.g., the age of the patient could be inferred from the density of the normal breast tissue and could be a useful predictor of malignancy. 

We cannot compare with papers focused on detection because our technique works on diagnosis of an already detected lesion \cite{mckinney2020international,wu2019deep}.

\FloatBarrier

\section{Discussion}

The high performance of uninterpretable models that appear to be leveraging mainly confounding information is a point of concern when incorporating models into clinical practice. 
Though a radiologist may not choose to view an explanation for every prediction, interpretable models still provide value over uninterpretable models. 
Because we know that AI systems fail \citep{zech2018variable}, we designed a system that can alert a radiologist to faulty reasoning at the time the prediction is made instead of only after the consequences of misprediction have been realized. The global interpretability (namely, the set of prototypes) allows the trained model to be fine-tuned by domain experts through pruning of prototypes that do not correspond to medically relevant features. The explanations provided can also be used for debugging a model and for retrospective analysis of model failures.

Our technique could be expanded with little change to include other BI-RADS features (e.g., mass shape). 
The technique might be able to be expanded to microcalcification clusters, the other main type of breast lesions, but there are more categories of calcification morphologies and the different types of cluster distributions can translate into lesions with extreme differences in scale which might pose interesting technical challenges. The underlying logic of the technique could be extended to digital breast tomosynthesis by representing a prototype as either a 2-dimensional part of a reconstructed slice image, or as a 3-dimensional portion of a tomosynthesis volume.

Future work with this model might include reader studies in which we measure any improvements in accuracy and radiologists report their trust in our system. Given the increased benefit of other AI assistance to less-experienced readers \citep{park2019computer,shimauchi2011evaluation}, it might be valuable to compare the benefit of this system to both sub-specialists and community radiologists who might be called on to do this work only occasionally. This work might help to extend the quality of care that patients receive at Duke (with highly-trained Duke radiologists) to patients that do not have access to this level of care.

The fine annotation techniques we developed to reduce the use of confounding information can be extended to other computer vision applications. The fine annotation technique could also be used on datasets with known confounders to see how effectively it reduces (or reveals) use of the confounders in its classification decisions.

\section{Conclusion}

Our work shows that we are able to create \textit{interpretable} mass margin prediction models with \textit{equal or higher performance} to their uninterpretable counterparts.
Using only a small dataset, we were able to provide an interpretable network that performs comparably with radiologists on mass margin classification and malignancy prediction.
The gradient stabilization improvement to the ProtoPNet training can be added into any future use of its codebase.

\FloatBarrier

\section{Author Contributions}

Idea and model development: A.J.B., F.S., C.T., C.C., J.L., C.R.. Code and code review: C.T., A.J.B., C.C.. Data collection: Y.R., A.J.B., F.S., J.L.. Data preprocessing: Y.R., C.T., A.J.B..

\section{Acknowledgements}

This study was supported in part by MIT Lincoln Laboratory, Duke TRIPODS and the Duke Incubation Fund.

We would like to acknowledge breast radiologists Michael Taylor-Cho MD, Lars Grimm MD, Connie Kim MD, and Sora Yoon MD, who annotated the dataset used in this paper. This study was supported in part by NIH/NCI U01-CA214183 and U2C-CA233254.

\FloatBarrier

\bibliography{main}

\FloatBarrier

\pagebreak
\appendix

\FloatBarrier
\section{Results Table}
\FloatBarrier

\begin{table}[h]
\small
  \caption{ \textit{Task:} The classification task. \textit{IAIA-BL:} The IAIA-BL model as described in Section \ref{sec:results_mal}. \textit{Radiologist Estimate:} The radiologist estimate as described in Section \ref{sec:results_mal}. \textit{VGG-16:} VGG-16 as described in Section \ref{sec:results_mal}.}
  \label{tab:mal_results}
  \centering
  \begin{tabular}{llll}
    \hline
    Task & \multicolumn{3}{c}{Test AUROC [95\% CI]}\\
     & IAIA-BL & Radiologist Estimate & VGG-16 \\
    \hline
    Malignancy (Test Set n=75) & 0.84 [0.74, 0.94] & 0.91 [0.85, 0.97] & 0.87 [0.82, 0.93] \\
    \hline
  \end{tabular}
\end{table}
\FloatBarrier
\section{Sample explanations} \label{app:local_anas}
\FloatBarrier

\begin{figure}[h]
        \begin{center}
            \includegraphics[width=\linewidth]{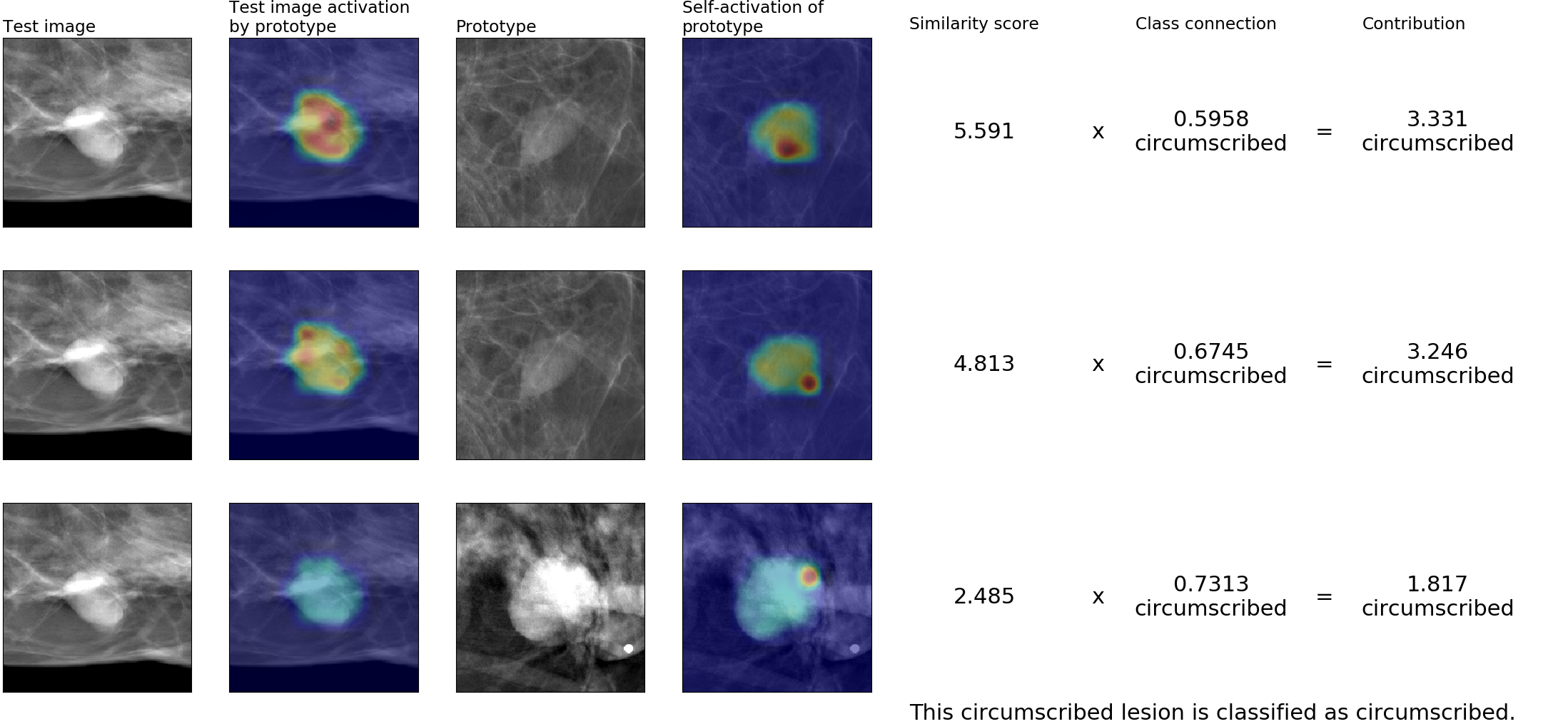}
        \end{center}
        \vfill
           \caption{This circumscribed lesion is correctly identified as circumscribed. The first two most activated prototypes are drawn from the same image, but are associated with different regions of that image.}
        \label{fig:auto_circ_p_circ}
\end{figure}

\begin{figure}[h]
        \begin{center}
            \includegraphics[width=\linewidth]{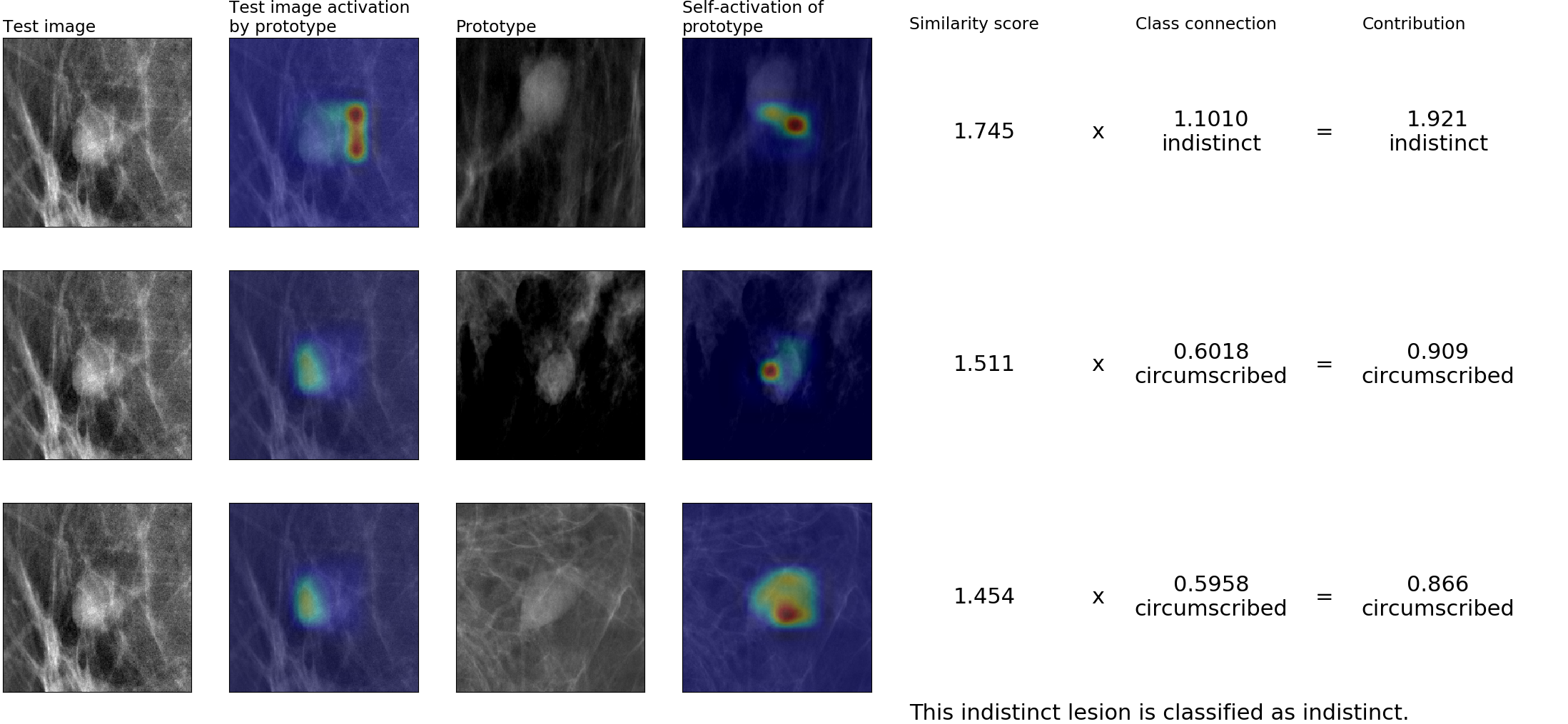}
        \end{center}
        \vfill
           \caption{This indistinct lesion is correctly identified as indistinct. The indistinct portion of the lesion margin (right side) activates the indistinct prototype and the circumscribed portion of the lesion margin (left side) activates the circumscribed prototypes.}
        \label{fig:auto_ind_p_ind}
\end{figure}

\begin{figure}[h]
        \begin{center}
            \includegraphics[width=\linewidth]{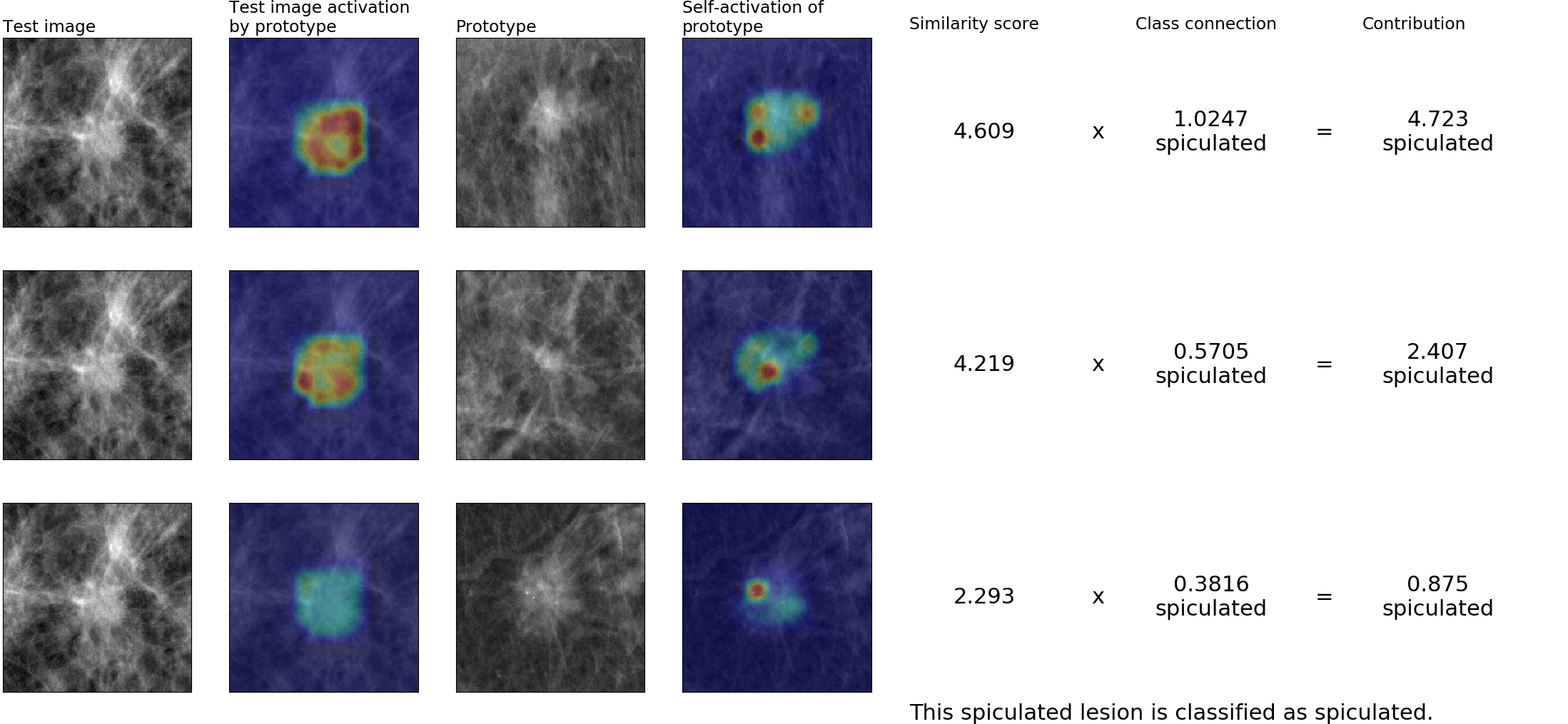}
        \end{center}
        \vfill
           \caption{This spiculated lesion is correctly identified as spiculated.}
        \label{fig:auto_spic_p_spic_2}
\end{figure}

\begin{figure}[h]
        \begin{center}
            \includegraphics[width=\linewidth]{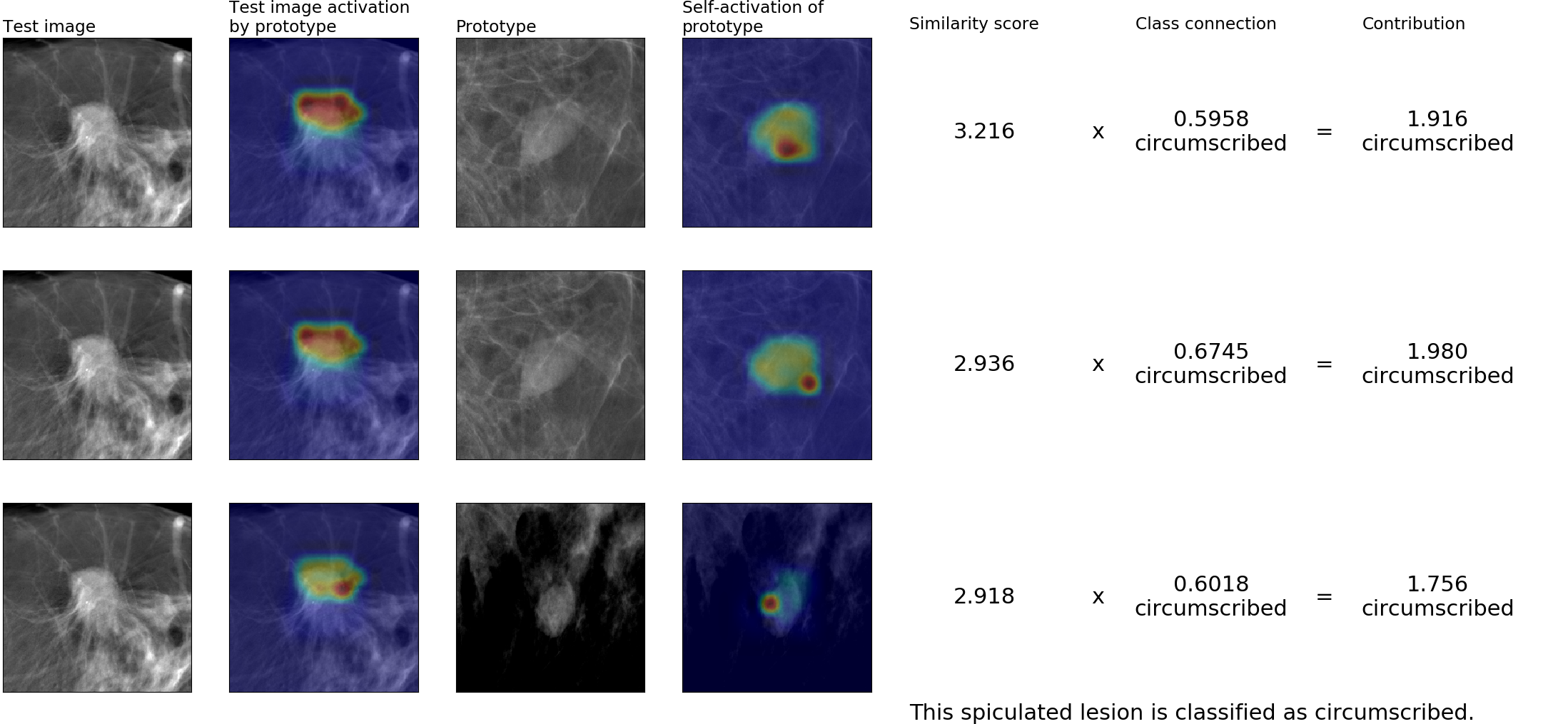}
        \end{center}
        \vfill
           \caption{This spiculated lesion is incorrectly identified as circumscribed. The explanation highlights only the circumscribed portion of the mass margin (top), but does not detect the spiculated portion (bottom).}
        \label{fig:auto_spic_pred_circ}
\end{figure}

Figures \ref{fig:auto_circ_p_circ}, \ref{fig:auto_ind_p_ind}, \ref{fig:auto_spic_p_spic_2} and \ref{fig:auto_spic_pred_circ} show explanations of mass margin classification automatically generated by IAIA-BL.


\FloatBarrier
\section{Comparing Visual Explanations} \label{app:compare_viz_expls}
\FloatBarrier

The IAIA-BL and ProtoPNet class activation visualizations shown in Figure \ref{fig:compare_gradcam_viz_to_us} are produced by taking the weighted average of the prototype activations maps for every prototype in the correct class. The weight for each prototype is the similarity score between the prototype and the original image. $\textrm{CAV}^{a,b}$, the value of the $a$-th row and $b$-th column of class activation visualization $\textrm{CAV}$, is defined as:

\begin{align}
    \textrm{CAV}^{a,b}(\mathbf{x}_i, y_i^{\text{margin}}) \backsim & \sum_{j: \text{class}(\mathbf{p}_j) ) = y^{\text{margin}}_i} s_j \left(\textrm{Upsample}\left(g_{\mathbf{p}_j}(f(\mathbf{x}_i))\right)\right)^{a,b}
\end{align}

where $\textrm{Upsample}\left(g_{\mathbf{p}_j}(f(\mathbf{x}_i))\right)$ is the prototype activation map for prototype $p_j$ on image $\mathbf{x}_i$. $\textrm{CAV}$ will have the same dimension as $\textrm{Upsample}\left(g_{\mathbf{p}_j}(f(\mathbf{x}_i))\right)$. $\textrm{CAV}$ is normalized using mix-max normalization so that its values fall between 0 and 1.

Compared to the class activation visualizations produced by baselines with similar predictive performance, the prototype activation maps produced by IAIA-BL are more likely to highlight the lesion and more likely to highlight the relevant part of the mass margin. This is shown quantitatively  by the activation precision metric results from Section \ref{sec:results_margin}.

\FloatBarrier
\section{Learned prototypes} \label{app:prototype_sets}
\begin{figure}[h]
        \begin{center}
            \includegraphics[width=0.8\linewidth]{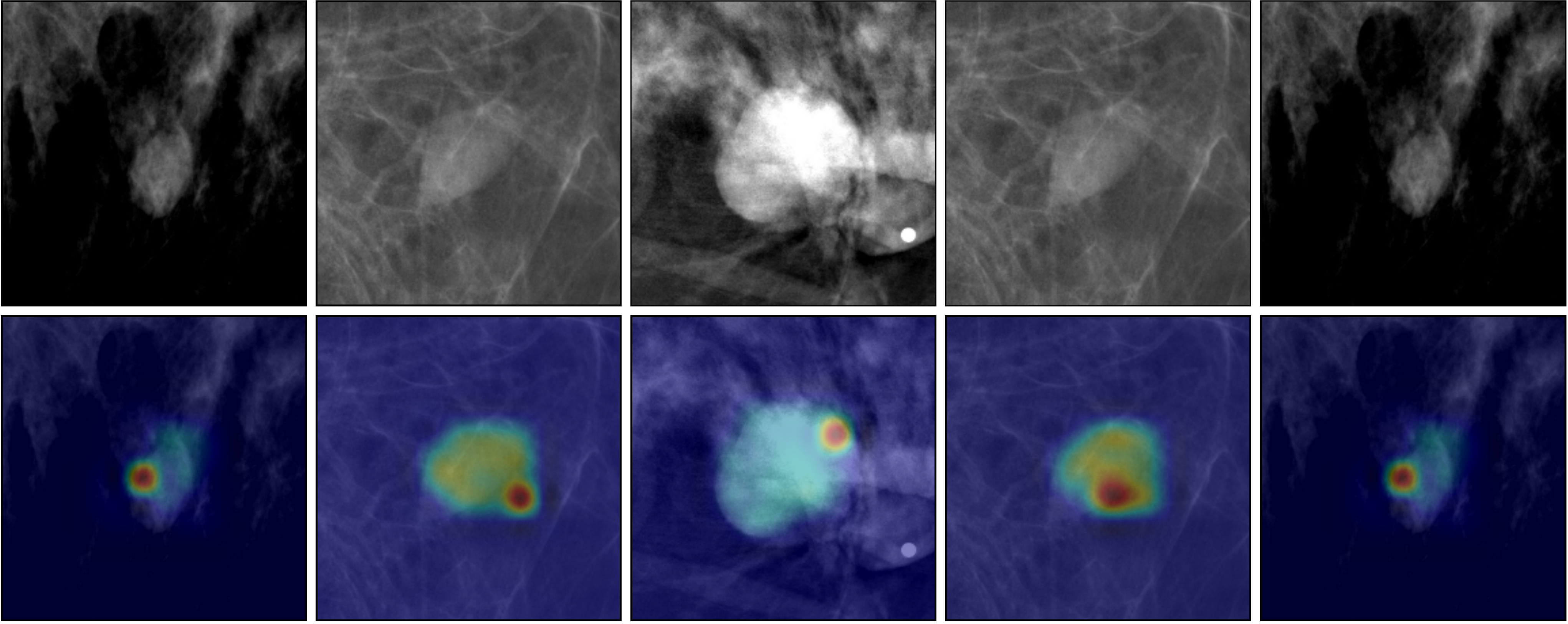}
        \end{center}
        \vfill
           \caption{Top: The learned prototypical circumscribed lesions from which the circumscribed prototypes (prototypes 0 to 4) are drawn. Bottom: The self activation of the prototypes (red indicates highest relevance part).}
        \label{fig:protos_pushed_on_all_circ}
\end{figure}

\begin{figure}[h]
        \begin{center}
            \includegraphics[width=0.80\linewidth]{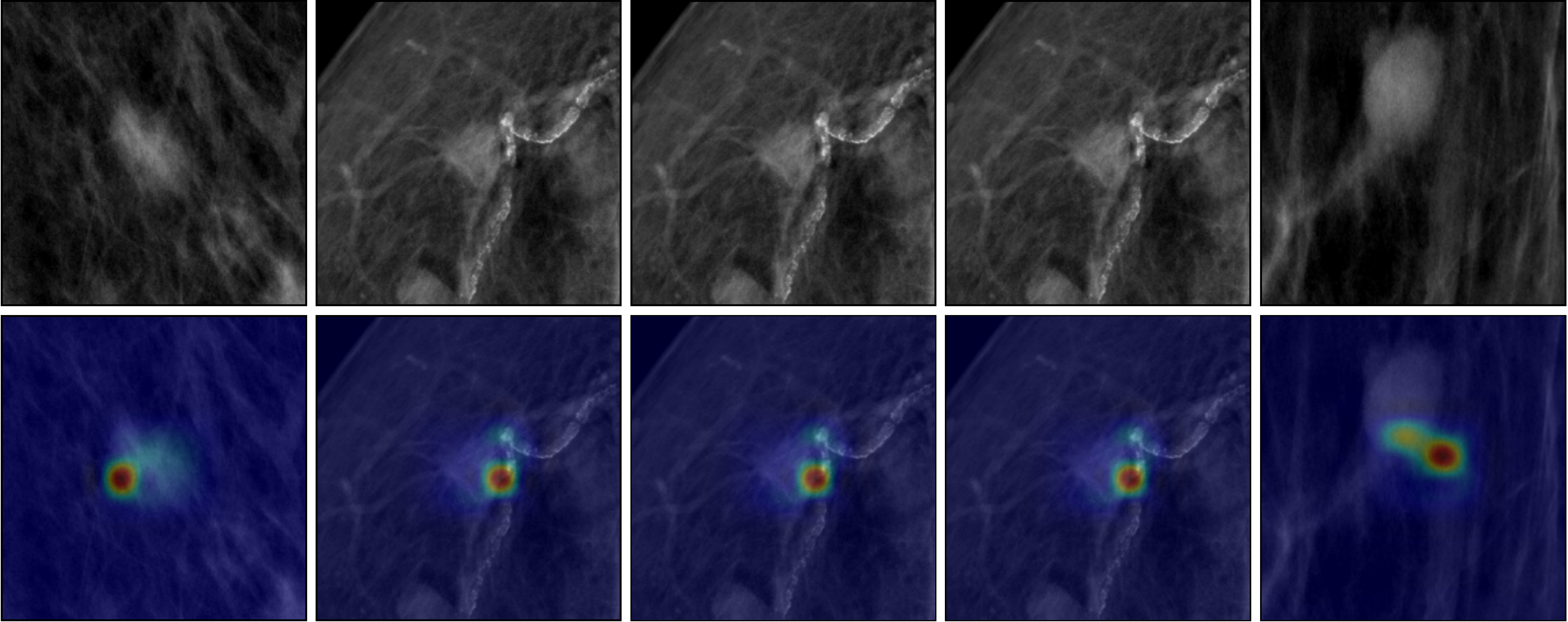}
        \end{center}
        \vfill
           \caption{Top: The learned prototypical indistinct lesions from which the indistinct prototypes (prototypes 5 to 9) are drawn. Bottom: The self activation of the prototypes (red indicates highest relevance part).}
        \label{fig:protos_pushed_on_all_ind}
\end{figure}

\begin{figure}[h]
        \begin{center}
            \includegraphics[width=0.80\linewidth]{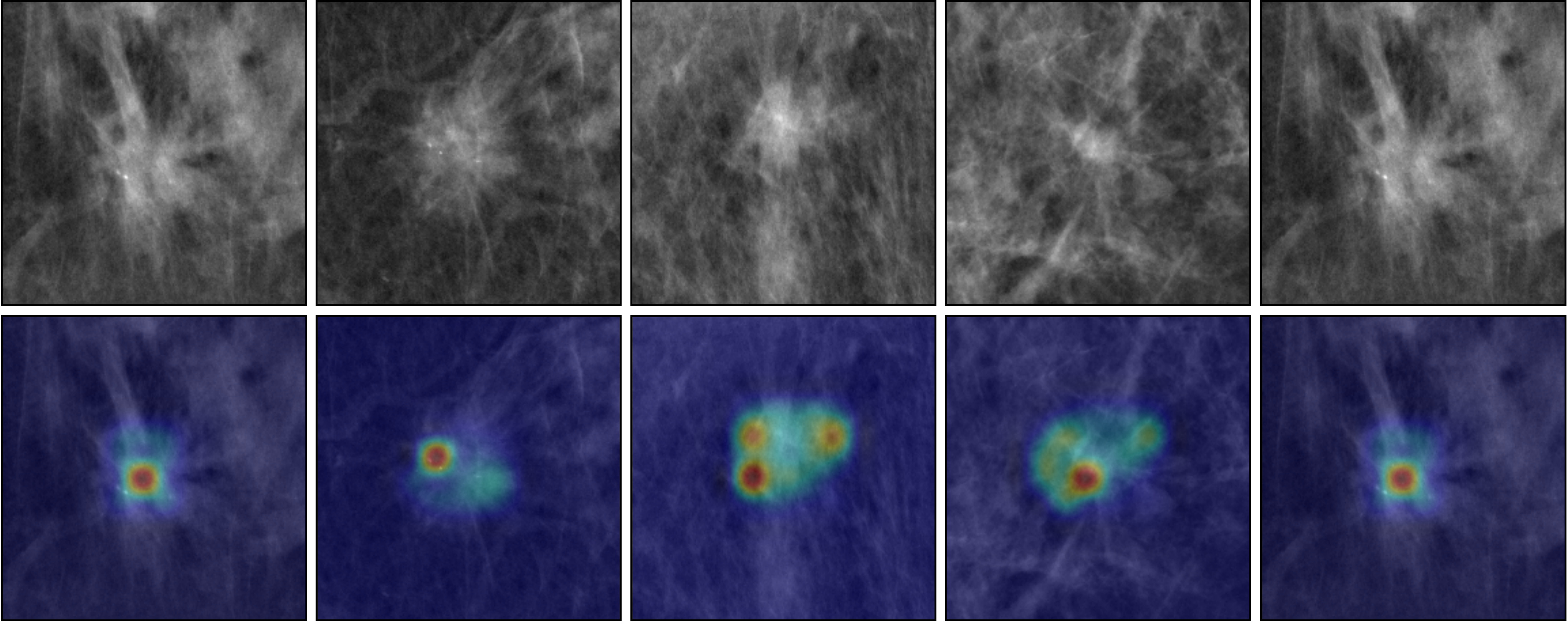}
        \end{center}
        \vfill
           \caption{Top: The learned prototypical spiculated lesions from which the spiculated prototypes (prototypes 10 to 14) are drawn. Bottom: The self activation of the prototypes (red indicates highest relevance part).}
        \label{fig:protos_pushed_on_all_spic}
\end{figure}

\FloatBarrier
Figures \ref{fig:protos_pushed_on_all_circ}, \ref{fig:protos_pushed_on_all_ind} and \ref{fig:protos_pushed_on_all_spic} show unpruned sets of prototypes learned by IAIA-BL. Duplicated prototypes 4, 6, 7 and 9 are pruned with a negligible loss in performance (0.001 decrease in AUROC).

\begin{figure}[h]
        \begin{center}
            \includegraphics[width=\linewidth]{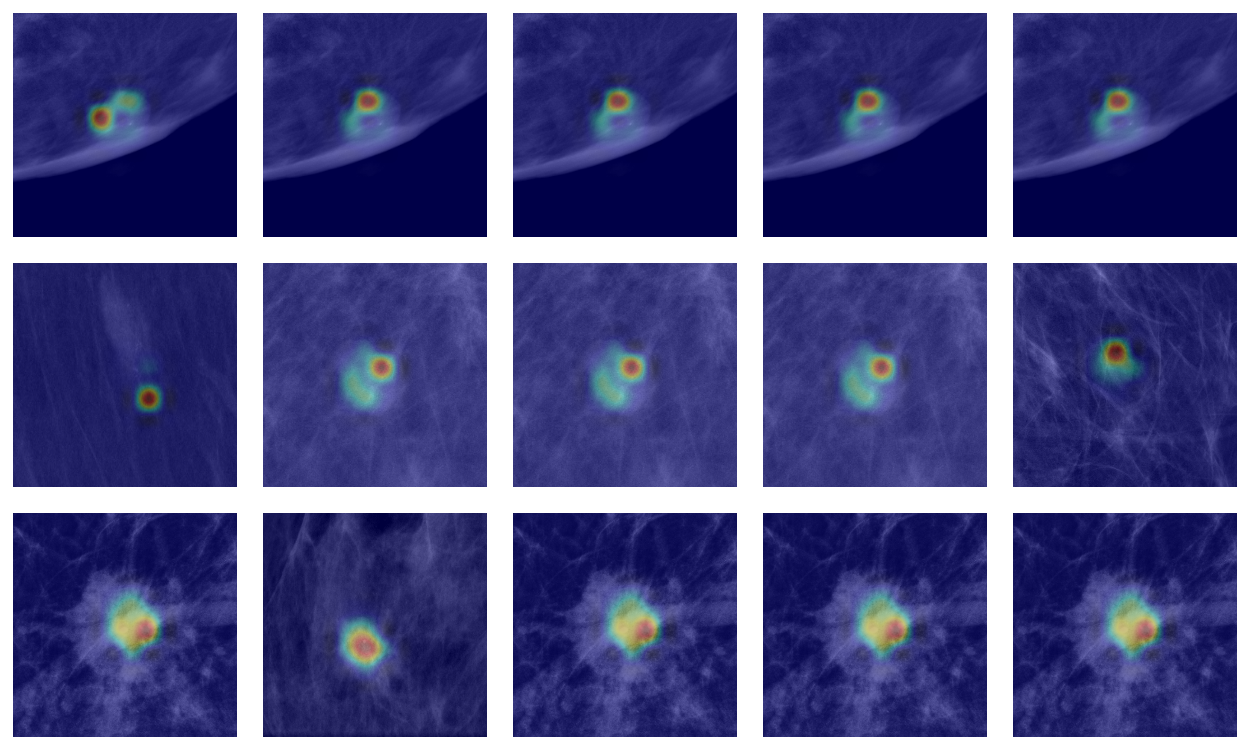}
        \end{center}
        \vfill
           \caption{This figure shows the self-activations for a set of prototypes that are trained by pushing onto only images with fine annotation, dataset $D'$. The rows from top to bottom correspond to circumscribed class, indistinct class and spiculated class. Duplicated prototypes 2, 3, 4, 7, 8, 12, 13 and 14 are pruned for being duplicates. Prototype 6 is pruned for having a higher circumscribed class connection than indistinct class connection even though its class is indistinct.}
        \label{fig:protos_pushed_on_fine}
\end{figure}

Figure \ref{fig:protos_pushed_on_fine} shows an unpruned set of prototypes learned by a variant of IAIA-BL that is constrained to select prototypes only from images with fine annotation, dataset $D'$. For this variant there is less prototype variety, but higher unpruned AUROC at 0.965 (compared to unpruned IAIA-BL at 0.955).

\FloatBarrier
\section{Hyperparameter tuning} \label{app:hyperparams}
\FloatBarrier

\begin{figure}[h]
    \centering
        \begin{subfigure}{0.4\textwidth}
        \includegraphics[width=\textwidth]{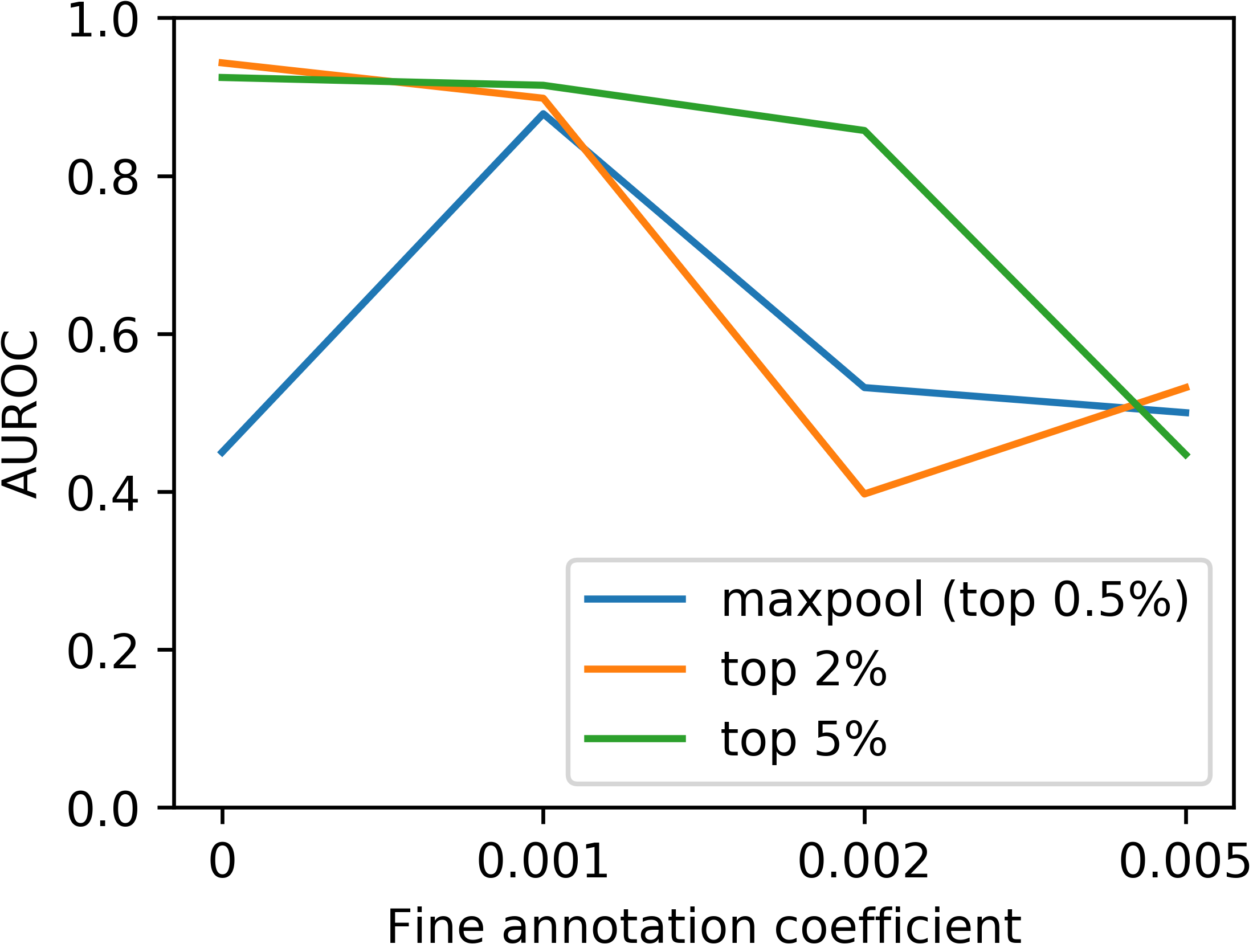}
        \caption{AUROC for varying fine annotation loss coefficient and top c\%, epoch 100.}
        \end{subfigure}
        \hfil
        \begin{subfigure}{0.4\textwidth}
        \includegraphics[width=\textwidth]{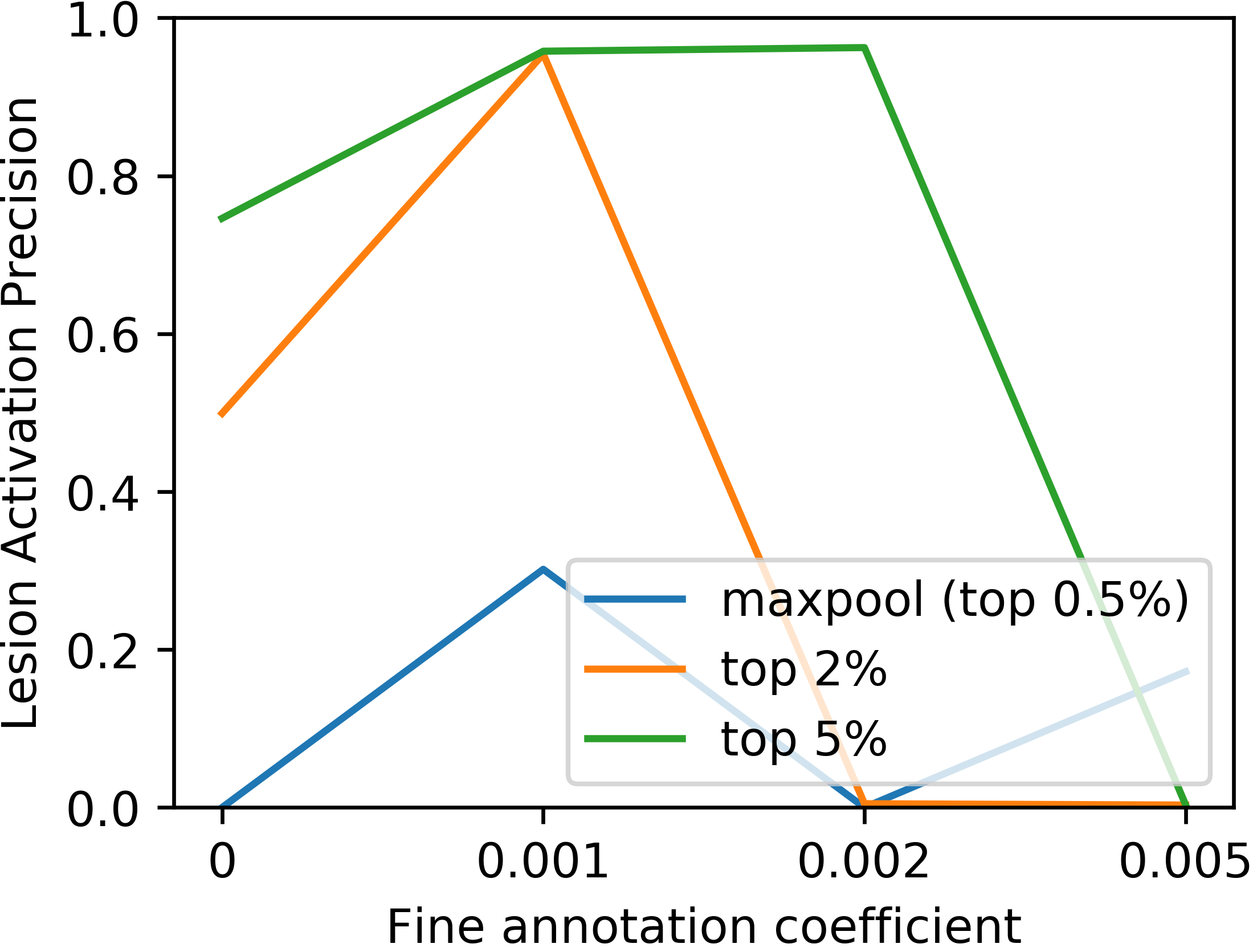}
        \caption{Activation precision for lesion-scale annotations, varying fine annotation loss coefficient and top c\%, epoch 100.}
        \end{subfigure}
        
    \caption{We explore how changes in hyperparameters affect key metrics of our model: performance (AUROC) and interpretability (activation precision). We show results for hyperparameter tuning over the following hyperparameters: fine annotation loss coefficient $\lambda_f$, and $c$ of top-$c$\%. These models were trained on the training set and tested on the validation set. Highest activation precision is found in the models with $\lambda_f=0.001$, $c=5$; and $\lambda_f=0.002$, $c=5$. Highest AUROC is found in the models with $\lambda_f=0.0$, $c=2$; $\lambda_f=0.0$, $c=5$; and $\lambda_f=0.001$, $c=5$. Models trained with $\lambda_f=0.0$ have lower interpretability than those trained with $\lambda_f=0.001$. Models trained with $\lambda_f\ge 0.002$ have lower performance than those trained with $\lambda_f=0.001$. For the model presented in the paper, we use $\lambda_f=0.001$, $c=5$, which was within the top hyperparameter combinations for both performance and interpretability as shown here. Note that ``maxpool'' logic (top 0.5\%) corresponds to the logic found in ProtoPNet \cite{PPNet}.}
    \label{fig:hyperparameters}
\end{figure}

For the IAIA-BL model presented in this paper, we use the hyperparameters: fine annotation coefficient 0.001 and top 5\% average pooling. These are selected based on hyperparameter tuning experiments trained on the training dataset and tested on the validation set, shown in Figure \ref{fig:hyperparameters}.
\FloatBarrier
\section{Context Window} \label{app:context}
\begin{figure}[h]
        \begin{center}
            \includegraphics[width=0.45\linewidth]{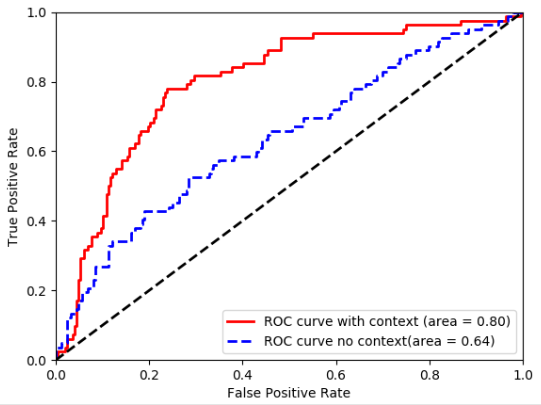}
          \vspace*{-20pt}
        \end{center}
        \vfill
          \caption{ROC curves for the task of spiculated vs$.$ other mass margins on CBIS-DDSM data. The results ``with context" are of the algorithm and test set that includes a 100 pixels on each side of the given ROI.}
        \label{fig:context}
\end{figure}

\FloatBarrier
While performing preliminary experiments with the publicly available dataset CBIS-DDSM, there is an option to download only the regions of interest (ROI). In preliminary work, we compared using only the ROI provided, and using the ROI provided but including an additional 100 pixels on each side of the ROI to provide context. Figure \ref{fig:context} shows that for the spiculated vs$.$ other mass margins task, the algorithm was able to achieve a AUROC of 0.80 on images that include context, but only 0.64 for the ROI images that do not. The reason for this may be that the provided boxes are too tight to convey the information needed for mass margin classification. This preliminary result informed our choice to include context around the ROIs provided by our radiologist annotations.

\FloatBarrier
\section{Figure for Activation Precision} \label{app:activation_precision}
\begin{figure}[h]
        \begin{center}
            \includegraphics[width=\linewidth]{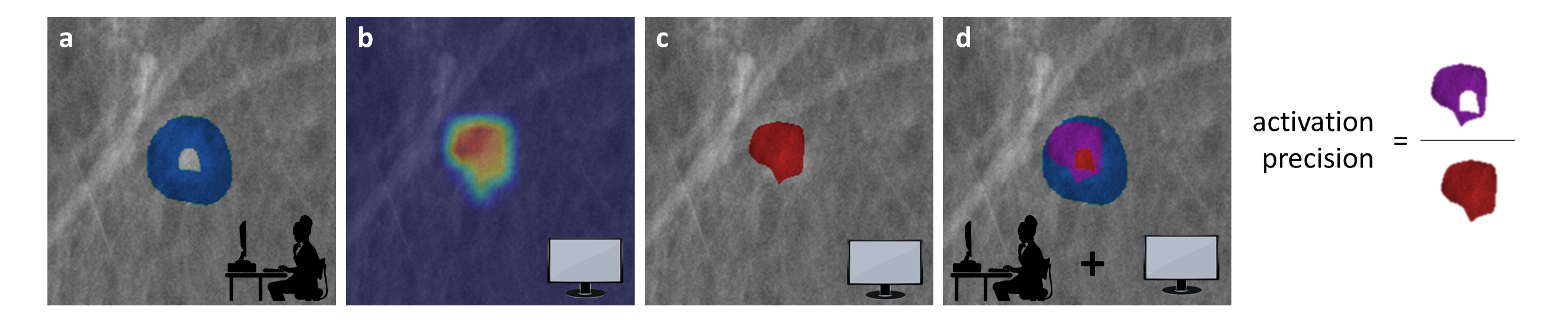}
          \vspace*{-20pt}
        \end{center}
        \vfill
          \caption{The definition for activation precision definition given as a visual example. (a) The doctor-annotated area of the image, indicating the medically relevant region for mass-margin classification, is highlighted in blue. (b) The activation of same-class prototype (learned in model training) on the lesion. The red region is most highly activated. (c) A mask of the top activation from \textit{b}. (d) The area of fine annotation from the radiologist annotator is shown in blue as in \textit{a}, the area most activated by a same-class prototype is shown in red as in \textit{c}, the overlap in these two regions is shown in purple.}
        \label{fig:ap_def}
\end{figure}

\FloatBarrier
A visual definition for activation precision is in Figure \ref{fig:ap_def}.

\FloatBarrier
\section{Expanded Discussion of Fine Annotation}
\label{app:fineannotation}
\FloatBarrier

Despite the promise of an interpretable mass margin classifier promised by a ProtoPNet, we found that a direct application of the ProtoPNet model and the training procedure in \cite{PPNet} failed to find medically relevant prototypes of various mass margin types. This was primarily because there was no extra supervision for prototype learning during the training of a ProtoPNet, which means that the ProtoPNet was free to choose any region of any mammogram in the training set as a ``prototypical case'' of a specific mass margin type, even if the region it had chosen did not contain the mass margin or the breast lesion. For example, a ProtoPNet we trained using the techniques in \cite{PPNet} yielded spiculated mass margin prototypes that contain healthy tissue as in Figure \ref{fig:fine_attention}(b), and the ProtoPNet was using the similarity with healthy tissue to decide if a lesion has a spiculated mass margin. This is an example of how a machine learning model could use confounding information to make predictions. The use of confounding information is fairly common in complex black-box models such as neural networks, and is especially dangerous in black-box models used for high-stakes decisions because those models do not explain their reasoning processes, and hence it is not easy to know whether they rely on confounding information or not \citep{rudin2019stop}.

In contrast, IAIA-BL addresses the confounding issues in a ProtoPNet trained for mass margin classification, by integrating additional supervision during prototype learning. As discussed, we annotated a small number of the training mammograms, by highlighting the mass margin in each of those training mammograms, and turning the highlighting into a mask where $1$ indicates that the pixel is in the highlighted mass margin and $0$ otherwise. We introduced a \textit{fine-annotation} loss during training, which encourages IAIA-BL to choose prototypes that are similar to the highlighted mass margins from the annotated mammograms of the training set. The fine-annotation loss we used resembles the gender attention loss used in \cite{tang2020mitigating}. However, the gender attention loss is not class/gender specific, in the sense that there are no separate terms for gender attention loss for male and that for female. In contrast, our fine-annotation loss \textit{takes into account the class identity of each annotated training mammogram}, in the sense that it not only encourages the prototypes belonging to the class of an annotated training mammogram to be similar to only the highlighted mass margin, but also requires the prototypes belonging to other classes to be dissimilar to any part of the mammogram. With our fine-annotation loss imposed on a small subset of annotated training mammograms, IAIA-BL for mass margin classification is not only able to achieve an area under receiver operating characteristic curve (AUROC) of $0.951$, but also able to learn medically relevant prototypes that capture the essential features of various mass margin types.

Let us walk through the example of learning a spiculated prototype, which detects the presence of spicules in the lesion margins. Suppose that we are interested in creating a network that predicts the presence of spiculated margin. To explain why an unknown lesion is identified as spiculated, we create explanations of the form ``this looks like that'' where the spicules of the unknown lesion look like the spicules of the prototypical lesion. When we train the network, we train for both \textit{which lesions become prototype lesions} as well as \textit{which part of the prototypical lesion is the relevant part.} An ideal spiculated lesion prototype would be a lesion with a spiculated margin where the margin itself is the most relevant part of the prototype. To train for this behaviour, we introduce a loss term during training that penalizes any spiculated prototype that has the most relevant part of the prototype being on an area other than the spicules. This constraint has the effect of directing model attention away from areas of the image that do not contain spicules, and encourages the training algorithm to select prototypes from image regions with spicules. This constraint on model attention penalizes the model for using confounding information, that is, information other than the spicules.

\FloatBarrier
\section{Expanded IAIA-BL Model Architecture}
\label{app:architecture}
\FloatBarrier

We use a variation on the ProtoPNet architecture from \cite{PPNet} as the underlying architecture of our mass-margin classifier in IAIA-BL. Figure \ref{fig:architecture} gives an overview of our IAIA-BL model. Given a region of interest $\mathbf{x}$ in a mammogram, our IAIA-BL model first extracts useful features $f(\mathbf{x})$ for mass-margin classification, using a series of convolutional layers $f$. In our experiments, the convolutional layers $f$ in our IAIA-BL model consist of all convolutional and max-pooling layers, \textit{excluding} the last max-pooling layer, in a VGG-16 network \citep{simonyan2015very}, followed by two additional $1 \times 1$ convolutional layers each with $512$ filters. We removed the last max-pooling layer from the base VGG-16, so that the resulting convolutional feature maps $f(\mathbf{x})$ will have a larger spatial dimension and a finer resolution. Since we resized all input images to $224 \times 224$, the spatial dimension of the convolutional feature maps is $14 \times 14$ (instead of $7 \times 7$ if the last max-pooling in the base VGG-16 were retained). The higher-resolution convolutional feature maps are able to represent finer details, such as variations in mass margin appearance, and are better suited for mass margin classification, because the mass margin is a relatively small part of the input image. Since the last convolutional layer has $512$ filters, there are $512$ convolutional feature maps in $f(\mathbf{x})$. In addition, the last convolutional layer also uses a sigmoid activation function, so that the convolutional features all lie between $0$ and $1$.

Like \cite{PPNet}, IAIA-BL has a prototype layer $g$, which follows the convolutional layers $f$. The prototype layer $g$ contains $m$ prototypes $\mathbf{P}=\{\mathbf{p}_j\}_{j=1}^m$ learned from the training set. In our experiments, each prototype is a $1 \times 1$ patch with the same number (i.e., $512$) of channels as the convolutional feature maps $f(\mathbf{x})$. Since each prototype has the same number of channels but a smaller spatial dimension than the convolutional feature maps, we can interpret each prototype as a prototypical activation pattern found among patches of convolutional feature maps, which, in our case, represents some prototypical feature associated with a specific mass margin type. For example, our IAIA-BL model learns prototypical representations of spicules for the spiculated mass-margin type, prototypical representations of fuzzy borders for indistinct mass-margin type, and prototypical representations of clearly defined borders for circumscribed mass-margin type, and stores these prototypical representations as prototypes in the prototype layer $g$ for later comparison. Given an input image $\mathbf{x}$, the prototype layer $g$ compares the input image $\mathbf{x}$ with each prototype $\mathbf{p}_j$, by computing the squared $\ell_2$ distances $d_{j,l}$ between $\mathbf{p}_j$ and all $1 \times 1$ patches of the convolutional feature maps $f(\mathbf{x})$, and transforming the distances $d_{j,l}$ into similarity scores using:
\[
s_{j,l} = \log\frac{d_{j,l}+1}{d_{j,l}+\epsilon}
\]
where we have $d_{j,l} = \|\mathbf{p}_j-f(\mathbf{x})_l\|_2^2$, $l \in \{(1,1),...,(1,14),(2,1),...,(14,14)\}$ indexes the $1 \times 1$ patches of the $14 \times 14$ convolutional feature maps $f(\mathbf{x})$, and $f(\mathbf{x})_l$ is the $l$-th $1 \times 1$ patch of the convolutional feature maps $f(\mathbf{x})$. Since each patch of the convolutional feature maps $f(\mathbf{x})$ has a similarity score with each prototype $\mathbf{p}_j$, the similarity scores between patches of $f(\mathbf{x})$ and a prototype $\mathbf{p}_j$ can be organized spatially into a similarity map, denoted $[s_{j,l}]_{l=(1,1)}^{(14,14)}$, which can then be upsampled to the size of the input image to produce a prototype activation map that identifies which parts of the input image are similar to the learned prototype. This is shown by the ``Sim. maps'' column of Figure \ref{fig:architecture}.

Each similarity map $[s_{j,l}]_{l=(1,1)}^{(14,14)}$ between an input image and a prototype $\mathbf{p}_j$ is reduced to a single similarity score $s_j$, summarizing the degree of similarity between the input image and the learned prototype. Unlike \cite{PPNet} who used max-pooling to reduce each similarity map to a single similarity score, we used \textit{top-k average pooling}. This idea is first found in the CNN literature as a subsection of Kalchbrenner et al. \citep{kalchbrenner2014convolutional} where dynamic top-k average pooling is used for sentiment analysis. For a given similarity map $[s_{j,l}]_{l=(1,1)}^{(14,14)}$, top-$k$ average pooling of the similarity map finds the $k$ highest similarity scores from the map and computes the average of those $k$ similarity scores. We denote this operation using $s_j = \text{AVGPOOL}({\text{topk}}([s_{j,l}]; k))$ in our paper. Note that max-pooling is a special case of top-$k$ average pooling, by using $k=1$. The top-$k$ average pooling allows the model to consider similarity between multiple parts of the input image and a mass-margin prototype, so that the similarity score after top-$k$ average pooling can be interpreted as how strong a prototypical feature is present (on average) in the $k$ most activated parts of the input image (instead of in the most activated part of the input image as in max-pooling). We use top-$k$ average pooling in IAIA-BL because the margin is distributed between multiple parts of the input image, so it makes semantic sense. In the IAIA-BL, top-$5\%$ average pooling (i.e., $k=\lfloor5\%(14 \times 14)\rfloor=9$) is used to reduce each similarity map $[s_{j,l}]$ to a similarity score $s_j$. The similarity scores between an input image and the learned prototypes are illustrated in the ``similarity score'' column in Figure \ref{fig:architecture}.

In our IAIA-BL, we initially allocated $5$ prototypes for each of the mass-margin types represented in our dataset. We use $\text{class}(\mathbf{p}_j)$ to denote the class identity of a prototype.

Fully connected layer $h_1$ multiplies the vector of similarity scores $[s_1, ..., s_m]$ by a weight matrix to produce three output scores $\hat{y}^{\text{circumscribed}}$, $\hat{y}^{\text{indistinct}}$, and $\hat{y}^{\text{spiculated}}$; one for each margin type. These are (afterwards)  normalized using a softmax function to generate the probabilities that the mass margin in the input image belongs to each of the three mass-margin types. The second fully connected layer $h_2$ then combines the vector of (unnormalized) mass-margin scores $\hat{\mathbf{y}}^{\text{margin}}=[\hat{y}^{\text{circumscribed}}, \hat{y}^{\text{indistinct}}, \hat{y}^{\text{spiculated}}]$ into a final score of malignancy $\hat{y}^{\text{mal}}$, which can be passed into a logistic sigmoid function to produce a probability that the input image has a malignant breast cancer.

This architecture provides both local interpretability by explaining each prediction in terms of the similarity between a given input image and the learned prototypes, as in Figure \ref{fig:IAIA_expls}, and global interpretability in terms of the clustering structure of the latent feature space (where semantically similar convolutional feature patches are clustered around prototypes representing the same semantic concepts). The set of learned prototypes is provided in Appendix \ref{app:prototype_sets}.
\FloatBarrier
\section{Expanded IAIA-BL Training}

The training of IAIA-BL differs from that of ProtoPNet \citep{PPNet} in three major ways: (1) Our IAIA-BL was trained with a fine-annotation loss which penalizes prototype activations on medically irrelevant regions for the subset of data with fine-scale annotations. For images with only lesion-scale annotation, fine-annotation loss penalizes prototype activations on regions other than the lesion. (2) Our IAIA-BL considers the top $5\%$ of the most activated convolutional patches that are closest to each prototype, instead of only the top most activated patch (using just the top activated patch is equivalent to top 0.5\% in our implementations) as in ProtoPNet. (3) We include an additional fully connected layer to go from mass margin scores $\hat{\mathbf{y}}^{\text{margin}}$ to malignancy score $y^{\text{mal}}$ whose training is isolated from the rest of the network. 
Some of the training regime from ProtoPNet is explained here to provide context for our changes. 

We represent the dataset of $n$ training images $\mathbf{x}_i$, with mass-margin labels $y^{\text{margin}}_i$ and malignancy labels $y_i^{\text{mal}}$, as $D=\{(\mathbf{x}_i, y^{\text{margin}}_i, y_i^{\text{mal}})\}_{i=1}^n$. A small subset $D' \subseteq D$ in the training set comes with fine-scale annotations. For a training instance $(\mathbf{x}_i, y^{\text{margin}}_i, y_i^{\text{mal}}) \in D'$ that comes with doctor's annotations of where medically relevant information is in that training image, we define a fine-annotation mask $\textbf{m}_i$, such that $\textbf{m}_i$ takes the value $0$ at those pixels that are marked as ``relevant to mass margin identification'' by a radiologist, and takes the value $1$ at other pixels. Each fine-annotation mask $\textbf{m}_i$ has the same spatial dimensions (height and width) as the training image $\textbf{x}_i$.

The learnable parameters of our IAIA-BL includes: (1) the parameters of the convolutional layers $f$, collectively denoted by $\mathbf{\theta}_f$, (2) the prototypes $\mathbf{p}_1$, ..., $\mathbf{p}_m$ in the prototype layer $g$, (3) the parameters of the first fully connected layer $h_1$, collectively denoted by $\mathbf{\theta}_{h_1}$, and (4) the parameters of the second fully connected layer $h_2$, collectively denoted by $\mathbf{\theta}_{h_2}$. 

The training of IAIA-BL is divided into four stages: (A1) training of the convolutional layers $f$ and the prototype layer $g$; (A2) projection of prototypes; (A3) training of the first fully connected layer $h_1$ for predicting mass-margin types; and (B) training of the second fully connected layer $h_2$ for predicting malignancy probability. Stages A1, A2, and A3 are repeated until the training loss for predicting mass-margin types converges, then we move to stage 4. By training stage B after converging  mass margin classification, we ensure that the mass margin classifier is not biased by the malignancy labels.
    
\textbf{Stage A1:} In the first training stage, we aim to learn meaningful convolutional features that can be clustered around prototypes that activate on medically relevant part of a given mammogram. In particular, we want convolutional features that represent a particular mass-margin type to be clustered around a prototype of that particular mass-margin type, and to be far away from a prototype of other mass-margin types. As in \cite{PPNet}, we jointly optimize the parameters $\mathbf{\theta}_f$ of the convolutional layers $f$, and the prototypes $\mathbf{p}_1$, ..., $\mathbf{p}_m$ in the prototype layer $g$, while keeping the two fully connected layers $h_1$ and $h_2$ fixed. Our training loss differs from that in \cite{PPNet}. In particular, we minimize the following training loss:
\begin{align}
    \textrm{min}_{\mathbf{\theta}_f, \mathbf{p}_1, ..., \mathbf{p}_m} & \frac{1}{n} \sum_{i=1}^n \textrm{CrossEntropy}(h_1 \circ \textrm{AVGPOOL} \circ \textrm{topk} \circ g \circ f(\mathbf{x}_i), y^{\text{margin}}_i) \nonumber\\ +& \lambda_c \textrm{ClusterCost} + \lambda_s \textrm{SeparationCost} + \lambda_f \textrm{FineLoss}.
    \label{app:eq:joint_training_obj}
\end{align}
The cross-entropy loss in Equation (\ref{app:eq:joint_training_obj}) penalizes  misclassification of mass-margin types on the training data (using the subnetwork that excludes the second fully connected layer $h_2$). It also ensures that the learned convolutional features and the learned prototypes are relevant for predicting mass-margin types. The cluster cost is defined by:
\begin{equation}
\textrm{ClusterCost} = \frac{1}{n}\sum_{i=1}^{n} \min_{j:\text{class}(\mathbf{p}_j)=y^{\text{margin}}_i} \left(\frac{1}{k}\sum\mathrm{mink}_{\mathbf{z} \in \text{patches}(f(\mathbf{x}_i))}\left(\|\mathbf{z}-\mathbf{p}_j\|_2^2\right)\right), \label{app:eq:cluster_cost}
\end{equation}
where $\mathrm{mink}$ gives the $k$ smallest squared distances between the convolutional patches of a training image and the $j$-th prototype, and $\frac{1}{k}\sum\mathrm{mink}$ gives the average squared distance over the $k$ smallest distances between the convolutional patches of a training image and the $j$-th prototype. The minimization of the above cluster cost encourages every training image to have $k$ convolutional feature patches that are close to a prototype of the same mass-margin type. The separation cost is defined by:
\begin{equation}
\textrm{SeparationCost} = -\frac{1}{n}\sum_{i=1}^{n} \min_{j:\text{class}(\mathbf{p}_j) \neq y^{\text{margin}}_i} \left(\frac{1}{k}\sum\mathrm{mink}_{\mathbf{z} \in \text{patches}(f(\mathbf{x}_i))}\left(\|\mathbf{z}-\mathbf{p}_j\|_2^2\right)\right). \label{app:eq:separation_cost}
\end{equation}
The minimization of the separation cost encourages the average of the $k$ smallest squared distances, between the convolutional patches of a training image and a prototype not of the same class as the training image, to be large. This encourages convolutional feature clusters of different mass-margin types to separate.

The fine-annotation loss (FineLoss) is new to this paper, and has not been previously explored by \cite{PPNet}. The purpose of the fine-annotation loss is to penalize prototype activations on medically irrelevant regions of doctor-annotated training mammograms. The fine-annotation loss is defined by:
\begin{equation}
\textrm{FineLoss} = \sum_{i \in D'}  \left(\sum_{j: \text{class}(\mathbf{p}_j) = y^{\text{margin}}_i} \|\mathbf{m}_i \odot \textrm{Upsample}(g_{\mathbf{p}_j}(f(\mathbf{x}_i)))\|_2
+ \sum_{j: \text{class}(\mathbf{p}_j) \neq y^{\text{margin}}_i} \|g_{\mathbf{p}_j}(f(\mathbf{x}_i))\|_2\right)
\end{equation}
where $g_{\mathbf{p}_j}(f(\mathbf{x}_i))$ computes the similarity map between patches of the convolutional features $f(\mathbf{x}_i)$ and the $j$-th prototype $\mathbf{p}_j$, and $\textrm{Upsample}(g_{\mathbf{p}_j}(f(\mathbf{x}_i)))$ computes bilinear upsampling of the similarity map $g_{\mathbf{p}_j}(f(\mathbf{x}_i))$ to yield a prototype activation map of the same dimensions (height and width) as the fine-annotation mask. 

Since the fine-annotation mask $\mathbf{m}_i$ and the prototype activation map (denoted $\textrm{Upsample}(g_{\mathbf{p}_j}(f(\mathbf{x}_i)))$) have the same dimensions, we can compute a Hadamard (component-wise) product between them. For a given training instance $i \in D'$ that comes with a fine-annotation mask $\mathbf{m}_i$ and has a mass-margin type $y^{\text{margin}}_i$, and for a mass-margin prototype $\mathbf{p}_j$ with $\text{class}(\mathbf{p}_j) = y^{\text{margin}}_i$, the Hadamard product between $\mathbf{m}_i$ and the prototype activation map (denoted $\textrm{Upsample}(g_{\mathbf{p}_j}(f(\mathbf{x}_i)))$) gives a map that shows the prototype activations in the medically irrelevant regions of the training image (because the fine-annotation mask $\mathbf{m}_i$ takes the value $1$ at medically irrelevant pixels, and $0$ at medically relevant pixels). Hence, the first sum in the parentheses of the fine-annotation loss encourages the amount of prototype activations in medically irrelevant regions to be small, for those prototypes that are of the same class as the training image $i \in D'$. This, in turn, encourages the training algorithm to learn prototypes that encode medically relevant mass-margin features for the prototypes' designated classes. On the other hand, for a given training instance $i \in D'$, the second sum in the parentheses of the fine-annotation loss penalizes any amount of prototype activation for a mass-margin prototype $\mathbf{p}_j$ with $\text{class}(\mathbf{p}_j) \neq y^{\text{margin}}_i$. This encourages the training algorithm to learn prototypes that stay away from any features that could appear in classes that are not prototypes' designated classes, so that the prototypes of a particular class represent distinguishing features of that class.

To incorporate the training data with fine annotations into model training, we optimize the convolutional layers $f$ and the prototype layer $g$ by minimizing the training objective in Equation (\ref{app:eq:joint_training_obj}), using stochastic gradient descent with $75$ training examples with lesion-scale annotation and $10$ training examples with fine annotation. The fine-annotation loss on a lesion-scale annotation penalizes activation outside of area marked as the lesion, whereas the fine-annotation loss on a finely annotated image penalizes activation outside of the region ``relevant to mass margin classification'' as marked by the radiologist. 

We initialized IAIA-BL using the parameters of a VGG-16 pre-trained on ImageNet \cite{deng2009imagenet} for the base convolutional layers. The additional convolutional layers were initialized randomly using Kaiming normal initialization \cite{he2015delving}. The prototype layer was initialized randomly using the uniform distribution over a unit hypercube (because the convolutional features from the last convolutional layer all lie between $0$ and $1$). For the first ten training epochs of IAIA-BL, we warmed up those randomly initialized layers, by only optimizing the training objective in Equation (\ref{app:eq:joint_training_obj}) with respect to the parameters in the two additional convolutional layers and the prototypes in the prototype layer.

\textbf{Stage A2:} As in \cite{PPNet}, we project the prototypes $\textbf{p}_j$ onto the nearest convolutional feature patch from the training set $D$, of the same class as $\textbf{p}_j$:
\begin{equation}
        \textbf{p}_j \leftarrow \textrm{arg min}_{\mathbf{z} \in Z_j} \|\mathbf{z} - \textbf{p}_j\|_2\text{; where } Z_j = \{\mathbf{z}: \mathbf{z} \in \text{ patches}(f(\mathbf{x}_i)) \text{ s.t. }\textrm{class}(\textbf{p}_j)=y^{\text{margin}}_i \forall i \in D\}.
\end{equation}

The projection of each prototype onto a convolutional patch from a training image allows us to visualize a prototype $\mathbf{p}_j$, by cropping out the region of that training image corresponding to the highest activations from the prototype activation map obtained from comparing the convolutional features of that training image with the prototype $\mathbf{p}_j$ (see \cite{PPNet} for a detailed description of how a prototype can be visualized).

\textbf{Stage A3:} After the previous two training stages, the (medically relevant) convolutional features have been clustered around mass-margin prototypes that are identical to some (medically relevant) convolutional features from training images, which can be visualized in the original image space. In this stage, we fine-tune the first fully connected layer $h_1$ to further increase the accuracy in predicting mass-margin types. In particular, we fix the parameters $\mathbf{\theta}_f$ of the convolutional layers $f$ and the prototypes $\mathbf{p}_1$, ..., $\mathbf{p}_m$, and minimize the following training objective with respect to the parameters $\mathbf{\theta}_{h_1}$ of the first fully connected layer $h_1$:
\begin{equation}
\textrm{min}_{\mathbf{\theta}_{h_1}} \frac{1}{n} \sum_{i=1}^n \textrm{CrossEntropy}(h_1 \circ \textrm{AVGPOOL} \circ \textrm{topk} \circ g \circ f(\mathbf{x}_i), y^{\text{margin}}_i). \label{app:eq:h1_training}
\end{equation}

The first time we enter stage A3, we initialize connections in fully connected layer $h_1$ to 1 for prototypes that are positive for that mass margin, -1 otherwise.

\textbf{Stage B:} In this stage, we train the second fully connected layer $h_2$ for predicting malignancy probability, using a logistic regression model whose input is the (unnormalized) mass-margin scores produced by the first fully connect layer $h_1$, and whose output is the probability of malignancy. Training an additional layer $h_2$ to predict malignancy allows us to compare our interpretable mass-margin classifier with previous non-interpretable classifiers that only output a malignancy label/probability. To prevent the malignancy information from biasing the mass margin classification, we train the model in a modular style and it is not trained completely end-to-end in any stage.

\FloatBarrier

\end{document}